%% file: main.tex
\documentclass[10pt,twocolumn,letterpaper]{article}

\usepackage[pagenumbers]{cvpr}
\usepackage{multirow}
\usepackage{amssymb}
\usepackage{bbm}
\usepackage{pifont}
\usepackage{arydshln}
\newcommand{\fullShape}{\mathcal{X}}
\newcommand{\partialShape}{\mathcal{Y}}
\newcommand{\vertFull}{\mathbf{V}_{\fullShape}}
\newcommand{\facesFull}{\mathbf{T}_{\fullShape}}
\newcommand{\vertPartial}{\mathbf{V}_{\partialShape}}
\newcommand{\facesPartial}{\mathbf{T}_{\partialShape}}
\newcommand{\gtCorres}{\mathbf{\Pi}_{\partialShape\fullShape}}
\newcommand{\ith}{i}
\newcommand{\cameraIth}{c^\ith}
\newcommand{\cameraSet}{\mathcal{C}}
\newcommand{\cameraFirst}{c^1}
\newcommand{\maxCamera}{N}
\newcommand{\cameraLast}{c^\maxCamera}
\newcommand{\imageFull}{\mathbf{I}^{\ith}_\fullShape}
\newcommand{\imageFeaturesFull}{\mathbf{Q}^{\ith}_\fullShape}
\newcommand{\imageFeaturesPartial}{\mathbf{Q}^{\ith}_\partialShape}
\newcommand{\shapeFeaturesFull}{\mathbf{F}_\fullShape}
\newcommand{\shapeFeaturesPartial}{\mathbf{F}_\partialShape}
\newcommand{\geoMatrix}{\mathbf{D}_\fullShape}

\newcommand{\shrec}{\textsc{Shrec16}}
\newcommand{\tosca}{\textsc{Tosca}}
\newcommand{\cuts}{\textsc{Cuts}}
\newcommand{\holes}{\textsc{Holes}}
\newcommand{\cutsTwentyFour}{\textsc{Cuts24}}
\newcommand{\cpTwopTwentyFour}{\textsc{CP2P24}}
\newcommand{\pfaust}{\textsc{Pfaust}}
\newcommand{\faust}{\textsc{Faust}}
\newcommand{\pfaustM}{\textsc{Pfaust-M}}
\newcommand{\pfaustH}{\textsc{Pfaust-H}}
\newcommand{\psmal}{\textsc{Psmal}}
\newcommand{\smal}{\textsc{Smal}}
\newcommand{\becos}{\textsc{BeCoS}}
\newcommand{\kids}{\textsc{Kids}}
\newcommand{\shrecTwenty}{\textsc{Shrec20}}
\newcommand{\dtFourd}{\textsc{DT4D}}
\newcommand{\scape}{\textsc{Scape}}
\newcommand{\dinoTwo}{DINOv2}
\newcommand{\dinoThree}{DINOv3}

\newcommand{\ciralitySym}{\text{chi}}
\newcommand{\acc}{\text{acc}}

\definecolor{cvprblue}{rgb}{0.21,0.49,0.74}
\usepackage[pagebackref,breaklinks,colorlinks,allcolors=cvprblue]{hyperref}

\def\paperID{****} 
\def\confName{CVPR}
\def\confYear{2026}

\title{Teaching DINOv3 About Partial 3D Geometry: \\A Self-Supervised Geometry-Aware Approach}

\newcommand{\authorspace}{\hspace{0.7cm}}
\newcommand{\affiliationspace}{\hspace{0.68cm}}
\author{
Viktoria Ehm$^{1,2}$
\authorspace
Dongliang Cao$^{3}$ 
\authorspace
Riccardo Marin$^{1,2}$
\authorspace
Daniel Scholz$^{1,2}$ \\
\authorspace
Weikang Wang$^{3}$
\authorspace
\authorspace
Florian Bernard$^{3}$
\authorspace
Daniel Cremers$^{1,2}$ \\
$^1$ Technical University of Munich 
\affiliationspace 
$^2$ Munich Center for Machine Learning
\affiliationspace
$^3$ University of Bonn
}

\begin{document}
\newcommand{\methodName}[0]{GeoLoRA}
\twocolumn[{%
\renewcommand\twocolumn[1][]{#1}%
\maketitle
\begin{center}
    \captionsetup{type=figure}
    \hspace{0.8cm}\includegraphics[width=0.82\linewidth]{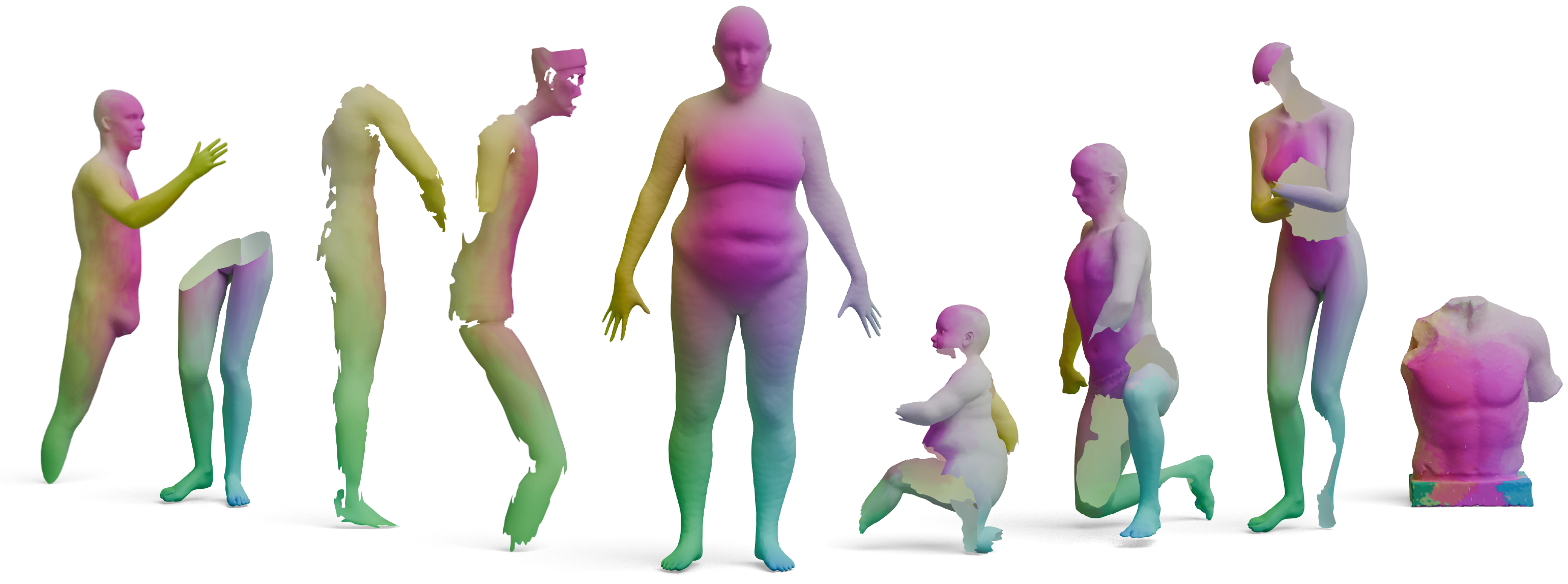}
    \caption{
    \textbf{\methodName{} (Geometry-Aware Low-Rank Adaptation)} injects geometry awareness into 2D foundation features for partial shape matching. Our method is self-supervised and produces robust features for different kinds and amounts of partiality, even for out-of-distribution real-world shapes (in the image, from left: two shapes cut by a plane; two partial views; a full template for reference; three shapes with holes; a statue from \cite{ScanTheWorldMMF}).
    }
    \label{fig:teaser}
\end{center}
}]

\input{sec/0_abstract}    
\input{sec/1_intro}
\input{sec/2_related_work}
\input{sec/3_method}
\input{sec/4_experiments}
\input{sec/5_conclusion}
\input{sec/6_acknowledgments}
{
    \small
    \bibliographystyle{ieeenat_fullname}
    \bibliography{main}
}

\input{sec/X_suppl}

\end{document}

%% file: sec/0_abstract.tex
\begin{abstract}
Partial shape matching is a crucial yet underexplored problem in 3D vision, with significant relevance to real-world scenarios where shapes are often only partially observed. Existing feature descriptors face difficulties in this setting, as traditional representations either struggle with the boundaries of partial shapes or heavily depend on the shape's spatial position. While existing approaches have employed DINO features for partial shape matching, these features are not inherently suited for handling partial observations. In this work, we propose a method to refine DINO features using LoRA-based self-supervised learning, enabling the generation of feature descriptors that are robust to partiality. Our features substantially improve performance on partial shape matching compared to traditional or vision foundation features. Additionally, when integrated into existing partial shape matching pipelines, we achieve state-of-the-art results on partial shape matching and left-right prediction benchmarks.
Our code is available at \url{https://github.com/vikiehm/geo-lora}.
\end{abstract}

%% file: sec/1_intro.tex
\section{Introduction}
\label{sec:intro}

Finding dense correspondences between 3D deformable shapes is a fundamental step in understanding geometric data, enabling several downstream tasks, including information transfer and learning statistical models.
Specifically, when the geometries do not share the same identity (e.g., different humans) and undergo non‑rigid deformations (e.g., different poses), establishing such correspondences remains an open research problem.
The challenge is further exacerbated when one (or both) of the shapes comes from partial observations, a common real-world case due to occlusions, acquisition noise, or local errors in the 3D reconstruction.

Scholars have relied on handcrafted descriptors~\cite{sun_concise_2009,aubry_wave_2011, tombari2010unique} for decades, both standalone and as input to deep learning pipelines.
Such descriptors demonstrate powerful capabilities for full shapes, but they often rely on theoretical premises (e.g., watertight meshes with the same topology) that make them impractical for partial use cases.
Recently, the emergence of foundation models like DINO~\cite{caron2021emerging, oquab2023dinov2, simeoni2025dinov3} or Stable Diffusion~\cite{rombach2022high} has introduced a new class of features.
Trained on billions of images in a self-supervised manner, these models demonstrate an impressive ability to characterise points semantically, while being robust across deformations and identities.
Such features have been applied to various computer vision tasks, including detection~\cite{liu2024grounding, zhang2022dino}, segmentation~\cite{li2023mask, wu2024dino}, and, more recently, 3D shape matching~\cite{dutt_diffusion_2024}.
However, these models have a limited understanding of 3D geometry, often exhibiting geometrical confusion, such as left-right mismatches and occlusions/disocclusions. 
These problems persist even in the most advanced foundation models (DINOv3~\cite{simeoni2025dinov3}), as we will demonstrate in our experiments. 
While recent works have explored 2D-based fine-tuning strategies~\cite{hartwig2025geco}, models trained solely on single-view images lack true 3D grounding.
Achieving this physical understanding requires incorporating 3D information directly into the training process.

While training a new foundation model for 3D shapes from scratch would require a huge amount of data and resources, we propose a more natural and elegant solution that injects 3D awareness into 2D foundation features.
We propose \textbf{\methodName{} (Geometry-Aware Low-Rank Adaptation)}, a new self-supervised approach that, using only synthetic data, informs the backbone 2D foundation features of the 3D structure.
Our idea is to start from full shapes and generate synthetic partial versions of them, using procedures to simulate real occlusions or artefacts.
Since the partial shapes originate from the full ones, they also enjoy a natural 3D correspondence.
We use this information to supervise a lightweight LoRA~\cite{hu2022lora} module on top of the 2D foundation model, rendering multiple images for both the full and partial shapes, and optimising the features to be aligned between the two.
Finally, unlike fully 2D approaches, since we have access to the 3D geometry of the shapes that generated the image, we incorporate a geodesic penalty into our contrastive loss, providing a strong geometric cue.
In practice, we learn a feature space where dissimilar points are pushed away proportionally to their distance on the object surface.

As a result, we achieve the best features for partial-to-full shape matching available to date.
We surpass handcrafted and 2D foundation feature alternatives across different partiality classes and datasets, both in partial-to-full and partial-to-partial settings (some results are shown in \cref{fig:teaser}).
The refined features integrate seamlessly with state-of-the-art matching methods, boosting their performance.
They also show better chirality (left-right disambiguation) than the original backbone features, suggesting that partial-to-full matching can be a useful auxiliary task for enhancing the geometrical understanding of foundation models.
In summary, our contributions are:
\begin{itemize}
\item We propose \textbf{\methodName{}}, a novel framework to inject 3D information into 2D foundation features for 3D partial observations.
\item We demonstrate that \methodName{} produces the \textbf{best features for partial 3D shapes} across multiple benchmark datasets.
\item We show that these features \textbf{improve SOTA performance} in downstream tasks as \textbf{partial-to-full and partial-to-partial shape matching}, and strengthen the model’s ability to handle geometrical challenges (e.g., left-right disambiguation).
\end{itemize}

%% file: sec/2_related_work.tex
\section{Related Work}
\label{sec:related_work}
In this section, we give an overview of the most relevant works, including partial shape matching, features for shape matching and self-supervised learning approaches.
\subsection{Partial Shape Matching}
While many full-to-full shape matching works use spatial features~\cite{ezuz2019elastic, eisenberger2020, Sundararaman2024deformation,cao2024spectral} or functional maps~\cite{ovsjanikov2012functional, litany2017deep, halimi2019unsupervised, cao_unsupervised_2023, bastian2023hybrid}, the task of matching partial observations remains a challenging open problem~\cite{ehm2025beyond}. We focus on the most important partial-shape matching works in this section and refer to these surveys~\cite{sahillioglu_recent_2020,deng_survey_2022} for an in-depth overview of general shape matching methods.

\paragraph{Partial-to-Full}
In partial-to-full shape matching, we aim to find correspondences between a partial observation and a full template shape.
PFM~\cite{rodola2017partial} analyses partial shape matching with functional maps and uses the slanted diagonal structure for optimisation.
DPFM~\cite{attaiki2021dpfm} incorporates this idea into a deep learning framework and adds a cross-attention refinement module for feature communication between the partial and full shape.
ULRSSM~\cite{cao_unsupervised_2023} ensures that the obtained functional map is coupled with valid point-wise correspondences.
Other partial shape matching methods focus on distance preservation~\cite{bracha2023partial, bracha_wormhole_2024} for isometric matching.
Axiomatic methods~\cite{ehm_geometrically_2024, roetzer2025geco, amrani2025highres} ensure smoothness through geometric consistency with an (Integer) Linear Programs.

\paragraph{Partial-to-Partial}
Partial-to-partial shape matching is even more challenging, as both shapes have missing parts that may also differ between the two.
In this setting, we not only want to find correspondences between shapes, but also determine the regions that overlap.
DPFM~\cite{attaiki2021dpfm} presents a supervised method, which predicts both overlap regions and correspondences.
Echo-Match~\cite{xie_echomatch_2025} improves DPFM's overlap predictor and shows the usefulness of 2D image foundation features for partial-to-partial shape matching.
Besides learning-based frameworks, GC-PPSM~\cite{ehm_partial--partial_2024} and Partial-GeCo~\cite{ehminteger} tackle the problem using a (non-)linear integer program, while enforcing geometric consistency.

\subsection{Features for Shape Matching}
\paragraph{Hand-Crafted Features}
Most full-to-full shape matching methods start from hand-crafted geometrical features, such as the Heat Kernel Signature~\cite{sun_concise_2009} and the Wave Kernel Signature~\cite{aubry_wave_2011}.
While showing success in full-to-full shape matching, partial shape matching features are fragile due to missing parts and boundaries. Hence, SHOT~\cite{tombari2010unique} features, or the vertex coordinates themselves, are generally preferred as input. On the other hand, their extrinsic nature leads to biases toward shape alignment, as shown in \cite{ehm2025beyond, xie_echomatch_2025, attaiki2023generalizable}.

\paragraph{Vision Foundation Models}
Vision foundation models (VFM), such as the DINO family~\cite{caron2021emerging,oquab2023dinov2,simeoni2025dinov3}, masked autoencoders~\cite{he2022masked}, and Segment Anything~\cite{kirillov2023segment} are trained on a large corpus of data providing general-purpose features for various tasks.
They also have proven useful in numerous domains outside of natural image analysis. 
Examples include adaptations in medicine for multi-modal images~\cite{scholz2025mm}, registration~\cite{song2024dinoreg}, and volume classification~\cite{muller2025medical}, as well as in robotics~\cite{liu2024grounding,di2024dinobot,zhou2024dino,wen2025data}. 
The field of shape matching has also leveraged these general-purpose features in seminal works on full~\cite{dutt_diffusion_2024,liu2025stable,wang2025kh} and partial~\cite{xie_echomatch_2025} shape matching.
Our work distinguishes itself from prior shape analysis papers by being the first to not only utilise the features generated by DINOv3 as input, but also incorporates 3D information about partial shapes into the foundation model.
However, it is worth mentioning that 2D foundation models suffer from geometrical ambiguity \cite{El_Banani_2024_CVPR, zhang2023tale}.
Several works rely on 2D tasks, incorporating geometrical augmentations such as image flipping \cite{Zhang_2024_CVPR}, or optimal transport objectives to account for (dis-)occlusions in the images \cite{hartwig2025geco}. 
Involving 3D data has been significantly less explored, and generally limited to textured rigid objects \cite{qian2025bridging}. 
To the best of our knowledge, we are the first to inject geometry awareness into foundation models for partial, non-rigid, untextured 3D shape matching.

\subsection{Self-Supervised Learning for 3D Shapes}
PointContrast~\cite{xie2020pointcontrast} introduces the contrastive loss called PointInfoNCE for pre-training on 3D point clouds, aiming to facilitate transfer learning to high-level scene understanding.
This approach was later extended by follow-up works~\cite{chen20224dcontrast, hou2021exploring}, all of which tackle static scenes, whereas we focus on non-rigid, deformed shapes.
In~\cite{attaiki2023generalizable}, the authors show that pre-training on rigid scenes can improve non-rigid shape matching performance by optimising the receptive field size of local features.
In~\cite{cao2023self}, they use a self-supervised learning approach together with the functional map framework to improve matching performance for shapes represented as point clouds.
While these methods show the effectiveness of self-supervised learning for 3D shapes, they do not yet tackle partial non-rigid shapes at a large scale using vision foundation models.

%% file: sec/3_method.tex
\section{Method}
\label{sec:method}

\begin{figure*}
    \centering
    \includegraphics[width=1\linewidth]{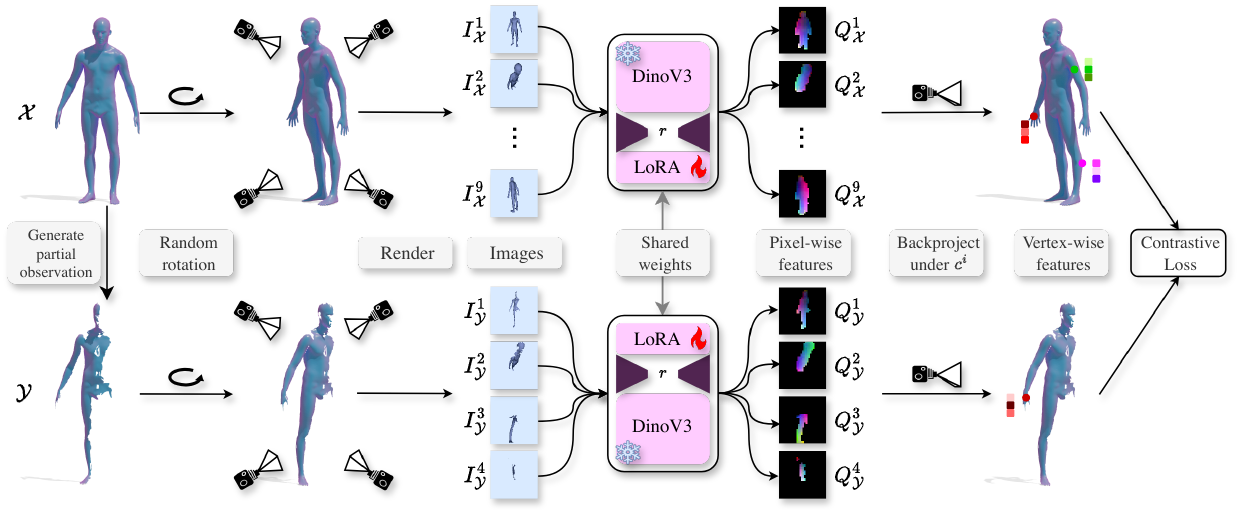}
    \caption{\textbf{Our Self-Supervised Training Pipeline:} Given a full mesh $\fullShape$,
    we first generate a partial observation $\partialShape$.
    After randomly rotating both shapes with different rotation matrices around the Y-axis, we render the partial and full shape from different perspectives with cameras $\cameraSet$.
    For every camera $\cameraIth \in \cameraSet$, we obtain an image $\imageFull$ of the full shape, which is then passed through a frozen DINOv3 architecture with a LoRA module with rank $r$ to extract pixel-wise features $\imageFeaturesFull$.
    These are then backprojected on the shape and averaged to get vertex-wise features $\shapeFeaturesFull$.
    Similarly, we proceed for the partial shape $\partialShape$.
    Finally, we use the Geodesic-Aware PointInfoNCE contrastive loss to align corresponding vertices of both shapes and separate all others based on geodesic distance.
    In our experiments we use $4$ views for the partial and $9$ views for the full shape.
    }
    \label{fig:pipeline}
\end{figure*}

\paragraph{Overview} 
\methodName{} is the first self-supervised approach that develops geometry-aware foundation features for partial shape matching.
We start by generating partial shapes from full ones, mimicking synthetic occlusions and missing surfaces.
Then, we use these pairs of full and partial shapes, and train a LoRA module to align the features between them, via a novel geodesic-based contrastive loss.
Figure~\ref{fig:pipeline} illustrates our approach, with individual components detailed below.
For better understandability, we refer to the supplementary material for a comprehensive outline of our notation.

\subsection{Self-Supervised Training}
\label{sec:selfsup}

\paragraph{Partial Shape Generation} 
Given a full 3D shape $\fullShape=(\vertFull,\facesFull)$ we generate a partial shape $\partialShape=(\vertPartial, \facesPartial)$ by randomly selecting one of the following approaches: for CUTS we follow~\cite{cosmo2016shrec, ehm_partial--partial_2024} and remove part of the shape using a random plane through its center; for HOLES we follow~\cite{bracha2023partial}, selecting a random number of areas with a random radius; for the RAY technique, we simulate the partiality induced by a partial observation from a random camera view~\cite{ehm2025beyond}.
These approaches generate partial shapes that are a subset of the full one ($\vertPartial \subset \vertFull$ and $\facesPartial \subset \facesFull$), which allows us to know the ground-truth correspondences $\gtCorres \in \mathbb{R}^{|\vertPartial|\times|\vertFull|}$ between them.

\paragraph{Foundation Features}
After the partial shape generation, we collect the foundation features for the full shape $\fullShape$ and its partial counterpart $\partialShape$.
We randomly rotate the shapes around the Y-axis, and following Diff3f~\cite{dutt_diffusion_2024} sample $\maxCamera$ different cameras around the full shape $\fullShape$, to obtain a set of cameras $\cameraSet = (\cameraFirst, ..., \cameraLast)$.
For each camera $\cameraIth$, we render an image $\imageFull \in \mathbb{R}^{512 \times 512 \times 3}$ from its viewpoint.
Each image is then processed by a DINOv3~\cite{simeoni2025dinov3} module.
Our goal is to improve the features obtained from the foundation model.
However, fine-tuning the entire foundation model would cause an additional computational burden.
Hence, we keep the DINOv3 backbone frozen during training and add a low-rank adaptation (LoRA) module with rank $r$ on top~\cite{hu2022lora}.
We use LoRA in a siamese fashion, meaning that the weights are shared between the full and partial observations.
For every image $\imageFull$ we obtain pixel-wise features $\imageFeaturesFull \in \mathbb{R}^{512 \times 512 \times 768}$, which we backproject to the mesh by averaging all pixels of all images observing a given vertex.
This produces vertex-wise features for the full shape, $\shapeFeaturesFull \in \mathbb{R}^{|\vertFull| \times 768}$.
We apply the same procedure to the partial shape $\partialShape$ by sampling $\maxCamera$ cameras around it, resulting in a set of cameras $\cameraSet$. For each camera $c_i \in \cameraSet$, we extract image features $\imageFeaturesPartial$, which we then backproject to produce vertex-wise features $\shapeFeaturesPartial \in \mathbb{R}^{|\vertPartial| \times 768}$.
At this point, it is worth highlighting that the full and partial shapes represent the same object in the same pose. 
At first glance, this may appear restrictive, as we actually aim to solve for non-rigid shape matching, and at inference time, we can expect to match objects that have undergone various deformations.
However, we believe our training setting exactly represents the task our backbone should solve: recover the semantics of an observed partial shape, as if it belonged to a full geometry.
In practice, we set a small number of $N=9$ cameras for the full shape and $N=4$ for its partial observation ensuring efficient model training.

\paragraph{Geodesic-Aware PointInfoNCE}
At this point, for every pair of full and partial shapes, we collected several rendered views and corresponding foundation features from the DINOv3 backbone and the appended LoRA module. Now, we aim to utilise this information to optimise the parameters of the LoRA module and enhance the alignment between full-shape and partial-shape features. For the loss, a common choice is the PointInfoNCE contrastive loss~\cite{xie2020pointcontrast, attaiki2021dpfm}, already shown promising results in previous shape matching works:

\begin{equation}
    L_{NCE} = -\sum_{(v_i, v_j) \in GT} \text{log} \frac{\text{exp}(\shapeFeaturesPartial^{i} \cdot \shapeFeaturesFull^{j} / \tau)}{\sum\limits_{v_k \in \vertPartial} \text{exp} (\shapeFeaturesPartial^{k} \cdot \shapeFeaturesFull^{j} / \tau) }, 
\end{equation}

where we define $GT$ as the set of corresponding pairs between the full and partial shapes, $\tau$ is a temperature hyperparameter, and $\shapeFeaturesFull$ and $\shapeFeaturesPartial$ are the features of the full and partial shapes, respectively. While the numerator pulls the feature representation of corresponding vertices of the partial and full shape together, the denominator pushes the features apart, making sure that the feature representation does not degenerate to one value.
However, in this formulation, there is a clear distinction between correct and incorrect predictions. Namely, it penalises all mismatches equally, without considering the severity of the error in the prediction. Hence, we add a weighting to the loss, such that points further away are penalised more heavily in the loss function, taking into account the actual distances between the vertices. 
We use the geodesic error matrix on the full shape, $\geoMatrix \in \mathbb{R}^{|\vertFull| \times |\vertFull|}$, where $\geoMatrix^{jk}$ represents the geodesic distance (distance along the surface) between two vertices $v_j, v_k \in \vertFull$.
When referring to a partial vertex $v_k \in \vertPartial$, we assume its known correspondence on the full shape, allowing us to index into $\geoMatrix$ appropriately.
We adjust the PointInfoNCE Loss, such that: 
\begin{equation}
    L={-}\sum_{(v_i, v_j) \in GT} \text{log} \frac{\text{exp}(\shapeFeaturesPartial^{i} \cdot\shapeFeaturesFull^{j} / \tau)}{\sum\limits_{v_k \in \vertPartial} \text{exp} ( (\geoMatrix^{jk}{+}0.5) {\cdot} \shapeFeaturesPartial^{k} {\cdot} \shapeFeaturesFull^{j} / \tau) }. 
\end{equation}
We shift the geodesic distance matrix by $0.5$ to obtain a similar range as in the original PointInfoNCE loss, where the denominator is multiplied by $1$.
Providing the network with this informative geometric cue produces strong supervision signals, which can provide fundamental characterisation of ambiguous geometry, such as symmetric points that are far apart (e.g., right vs. left limbs). 
Our intuition will be later confirmed in the experiments, where our method shows improvements over the vanilla PointInfoNCE for left-right disambiguation, and in terms of geodesic error.

%% file: sec/4_experiments.tex
\section{Experiments}
\label{sec:experiments}
\begin{table*}[htp]
\small
    \centering
    \begin{tabular}{llccccc|cc}
        \toprule
        Type & Dataset & WKS & HKS & Diff3f & \dinoTwo{} &  \dinoThree{} & Trained on & \textbf{\methodName}\\
        \midrule
        \multirow{5}{*}{\textbf{P2F}} & \becos{}~\cite{ehm2025beyond} & 58.35 & 69.50 &25.61 &21.54 & 19.95 & \becos{} & \textbf{5.57}
        \\
         &\shrec{} \cuts{}~\cite{cosmo2016shrec} &62.08 & 53.32 & 28.01 & 26.58 & 21.25 & \becos{}* &\textbf{10.86}
         \\
         &\shrec{} \holes{}~\cite{cosmo2016shrec} &50.33 & 65.56 & 26.67 & 20.22 & 20.22 & \becos{}* & \textbf{9.70}
         \\
         & \pfaustM{}~\cite{bracha2023partial} &50.28 & 52.19 & 12.60 & 11.38 & 11.36 &  \faust{} & \textbf{2.88}
         \\
         & \pfaustH{}~\cite{bracha2023partial} & 54.75 & 60.83 & 19.32 & 17.47 & 13.82 &  \faust{} & \textbf{3.33}
         \\
        \midrule
        \multirow{3}{*}{\textbf{P2P}} & \becos{}~\cite{ehm2025beyond} & 41.35 & 40.48 & 19.45 & 17.43 & 16.84 &  \becos{} & \textbf{6.72}
        \\
        & CP2P24~\cite{attaiki2021dpfm, ehm_partial--partial_2024} & 38.40 & 37.04 & 22.60 & 18.43 & 18.55 & \becos{}* & \textbf{12.46}
        \\
        & \psmal{}~\cite{ehm_partial--partial_2024} & 39.50 & 39.53 & 30.86 & 19.59 & 18.64 & \smal{} & \textbf{6.85}
        \\
        \bottomrule
        \end{tabular}
    \caption{\textbf{Improvement of Feature Quality Through Our Self-Supervised Pipeline:} We show that with \methodName{} we can improve the feature quality on all partial datasets in terms of the mean geodesic error ($\times 100$). 
    \becos{}* indicates \becos{} without \shrec{} data.
    }
    \label{tab:features_quantitative}
\end{table*}

We demonstrate the effectiveness of \methodName{} with an extensive set of experiments. First, we compare the matching induced by the features from \methodName{} with those obtained from handcrafted and learning-based alternatives.
Further, we incorporate \methodName{} features into partial-to-full and partial-to-partial shape matching SOTA methods, demonstrating their natural blend with them.
To show the improved geometrical representation, we also test their left-right characterisation on partial shapes.
Finally, we justify our design choices through an ablation study.

\subsection{Implementation Details}
We use the DINOv3 ViT-B/16 backbone and integrate LoRA modules into all attention layers. 
The LoRA configuration uses a rank of $r=16$ and a scaling factor of $\alpha=32$.
To maintain training feasibility, we use $9$ views for each full shape and $4$ views for each partial shape. During inference, we increase the number of views to $49$.
In each training batch, we generate two partial observations per full shape.
All methods are trained for $50.000$ iterations.

\subsection{Datasets}
For partial-to-full matching, we use the \textbf{\shrec{}} dataset \cite{cosmo2016shrec}, which is based on the synthetic shapes of \tosca{}~\cite{bronstein2008numerical}. It contains $76$ nearly-isometric shapes from eight different classes of humans and animals. Each class has one template mesh in a neutral pose.
The partial shapes are divided into \cuts{} and \holes{}.
The \cuts{} dataset contains partial shapes cut by a plane, whereas the \holes{} dataset contains shapes with irregular holes that are produced by an erosion process. We rely on the \cutsTwentyFour{} split~\cite{ehm_partial--partial_2024} to ensure that no shapes from the test set have been used for training. We also test on the \textbf{\pfaust{}} dataset \cite{bracha2023partial}, a modification of \cite{bogo2014faust}, where $10$ different humans ($8$ are used for training and $2$ for testing) are seen in $10$ different poses. \pfaust{} processes the dataset adding artificial holes of varying difficulty, including \pfaustM{} (medium difficulty) and \pfaustH{} (hard difficulty). A relevant dataset is also \textbf{\textsc{CP2P24}} \cite{attaiki2021dpfm, ehm_partial--partial_2024}, a partial-to-partial matching dataset based on \shrec{}, including cuts for near-isometric shapes. We also stress-test our method on non-isometric partial correspondence tasks using the \textbf{\psmal{}} dataset \cite{ehm_partial--partial_2024}, derived from \smal{} \cite{zuffi2017small} by adding artificial cuts. For \psmal{}, the train and test sets are split by different animal species. Finally, we also include the recent \textbf{\becos{}} dataset \cite{ehm2025beyond}, which integrates both humanoid and animal shapes from \faust{} \cite{bogo2014faust}, \scape{} \cite{anguelov2005scape}, \smal{} \cite{zuffi2017small}, \dtFourd{} \cite{li20214dcomplete}, \tosca{} \cite{bronstein2008numerical}, \kids{} \cite{rodola2014dense}, and \shrecTwenty{} \cite{dyke2020shrec}, totaling over 2,500 unique shapes. \becos{} provides the most extensive benchmark, with full-to-full, partial-to-full, and partial-to-partial shape instances.

\subsection{Metrics}
We report the widely used \textbf{Geodesic Error} metric following the Princeton Protocol~\cite{kim2011blended}.
The error is scaled by the shape diameter of the corresponding full shape, also for partial-to-partial shape matching~\cite{ehm_partial--partial_2024}.
We report the mean geodesic error over all vertices in the dataset.
For the partial-to-partial datasets, we report the geodesic error on the overlapping region. 
As proposed in~\cite{ehm_partial--partial_2024}, we use the \textbf{Intersection over Union}~(IoU) to evaluate the overlap region prediction.
For a shape $\partialShape$ we compare the overlap prediction $P_{\partialShape} \in \{0,1\}^{|\vertPartial|}$ and the ground truth overlap $G_{\partialShape} \in \{0,1\}^{|\vertPartial|}$, which includes for every vertex in $\vertPartial$ if it is in the overlapping region.
We evaluate the quality of the matching with the $\text{IoU} = \frac{|P_{\partialShape} \cap G_{\partialShape}|}{|P_{\partialShape} \cup G_{\partialShape}|}$.
Finally, we follow~\cite{wang2025kh} for the evaluation of \textbf{Left-Right (chirality) Accuracy}. For every partial shape $\partialShape$ with vertices $\vertPartial$, we get the predicted chirality vector $P_\ciralitySym \in \{0,1\}^{|\vertPartial|}$, where $0$ indicates one side and $1$ the other side.
We compute the accuracy with the ground truth chirality vector $G_\ciralitySym \in \{0,1\}^{|\vertPartial|}$ as:
\begin{equation}
    \mathrm{acc}_\partialShape = \frac{1}{|\vertPartial|}\sum_{v\in \vertPartial} \mathbbm{1} (\text{sign}(P_\ciralitySym)=G_\ciralitySym),
\end{equation}
where $\mathbbm{1}$ is the indicator function. 
We refer to the supplementary for the computation for a full dataset.

\subsection{Matching with Raw Features}
\label{sec:features_match}
\newcommand{\xfausth}{0.11}
\newcommand{\faustHFig}[1]{\includegraphics[width=\xfausth\linewidth]{figures/faust_h/#1.png}}
\newcommand{\faustHFigSource}[1]{\includegraphics[width=0.15\linewidth]{figures/faust_h/#1.png}}
\newcommand{\xcuts}{0.12}
\newcommand{\cutsFig}[1]{\includegraphics[width=\xcuts\linewidth]{figures/shrec16_cuts/#1.png}}
\newcommand{\xbecos}{0.15}
\newcommand{\becosPfFig}[1]{\includegraphics[width=\xbecos\linewidth]{figures/becos_p2f/#1.png}}

\begin{figure*}[htp!]
\centering
\setlength{\tabcolsep}{-7pt}
\begin{tabular}{lcccccccc}
\toprule
  & Source & GT & HKS & WKS & Diff3f & \dinoTwo{} & \dinoThree{} & \textbf{\methodName} \\
  \midrule
     \rotatebox{90}{\hspace{0.5cm}\becos{}} & \becosPfFig{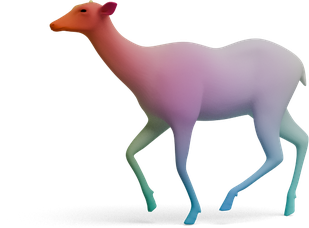} & \becosPfFig{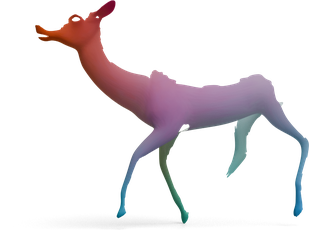} & \becosPfFig{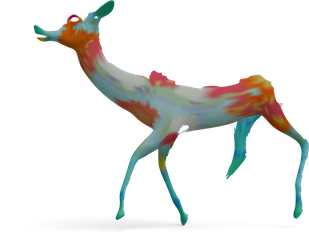} & \becosPfFig{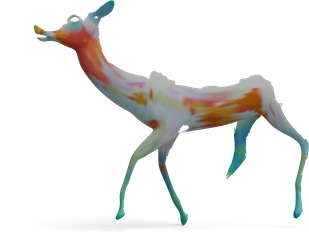} & \becosPfFig{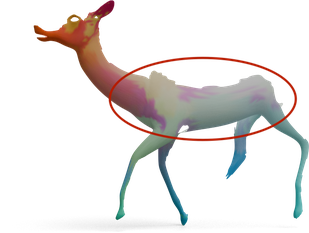} & \becosPfFig{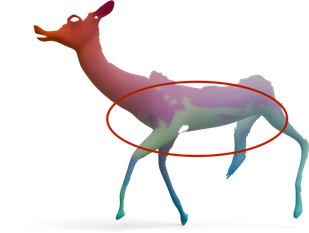} & \becosPfFig{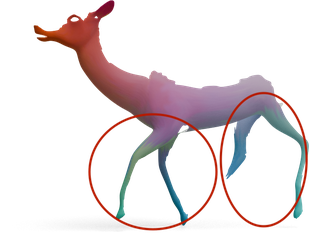} & \becosPfFig{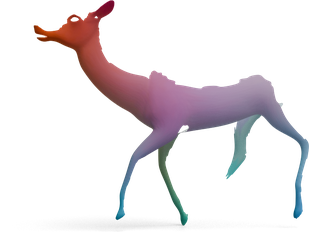} \\
    \rotatebox{90}{\hspace{0.4cm}\cuts{}} & \cutsFig{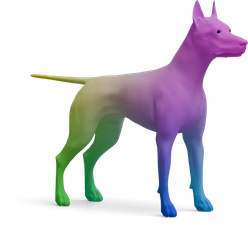} & \cutsFig{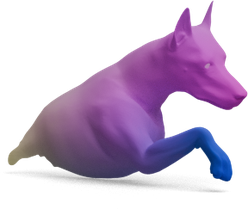} & \cutsFig{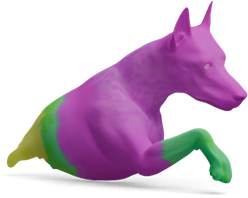} & \cutsFig{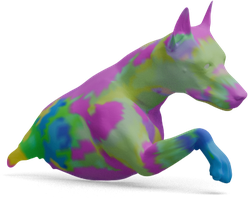} & \cutsFig{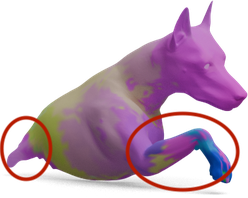} & \cutsFig{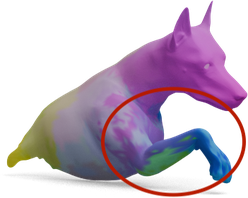} & \cutsFig{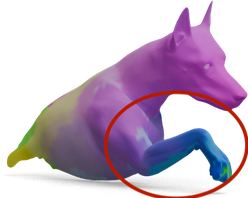} & \cutsFig{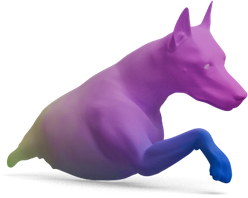} \\
   \rotatebox{90}{\hspace{1cm}\pfaustH{}} & \faustHFigSource{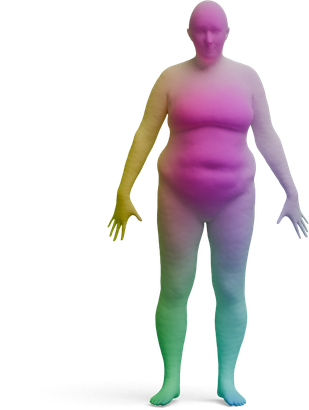} & \faustHFig{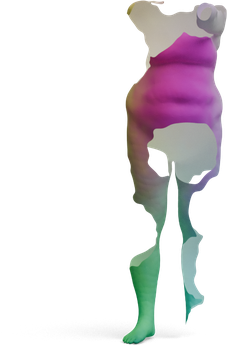} & \faustHFig{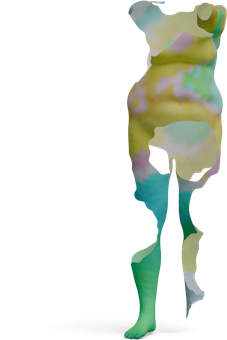} & \faustHFig{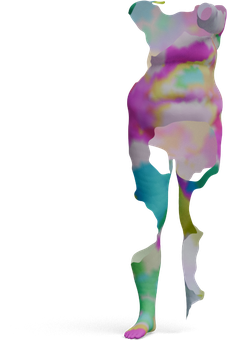} & \faustHFig{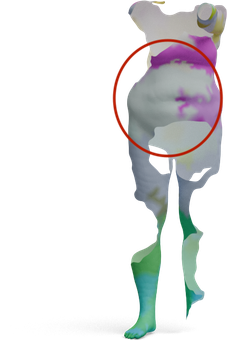} & \faustHFig{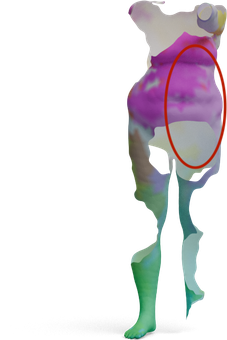} & \faustHFig{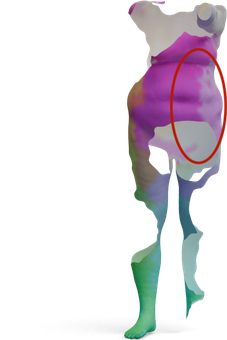} & \faustHFig{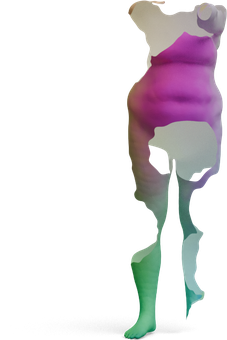}
   \\
   \bottomrule
\end{tabular}
\caption{\textbf{Qualitative Results of Feature Quality:} We show colour-coded correspondences based on the features computed with different feature extractors. \methodName{} exhibits improved performance, especially in left-right predictions.
}
\label{fig:qualitative_p2f}
\end{figure*}

\textbf{Baselines and Data.}
We compare matchings induced by \methodName{} with established handcrafted descriptors—WKS \cite{aubry_wave_2011} and HKS \cite{sun_concise_2009}—as well as features from foundation models \dinoTwo{} \cite{oquab2023dinov2}, \dinoThree{} \cite{oquab2023dinov2}, and Diff3f \cite{dutt_diffusion_2024}.
For DINOv2, DINOv3, and \methodName{}, we utilise 49 views, providing the optimal balance between performance and runtime (see our ablation study).
For training our self-supervised network, we always use the full shapes from the training set, ensuring that no shapes of the test/validation set have been seen during training.
Since \shrec{}'s training set includes full shapes of partials used in its test set, we create \becos{}*, the \becos{} training dataset excluding \shrec{} data.
We evaluate the geodesic error of these feature extractors by performing a nearest-neighbour search using the L2 distance.

\textbf{Results} We report the quantiative analysis in Table~\ref{tab:features_quantitative}. We observe a significant improvement in terms of geodesic error over all the datasets, particularly for the very challenging \becos{} dataset. We report qualitative examples in Figure~\ref{fig:qualitative_p2f}. Our improvement is noticeable across all datasets, with better left-right disambiguation compared to using \dinoThree{} features directly.

\subsection{Integration with Matching Methods}
State-of-the-art shape matching methods do not use features as they are, but further refine them for specific datasets and settings. Hence, we also explore the applicability of our features as replacements for alternatives in matching pipelines, both for the partial-to-full and partial-to-partial matching methods. Notice that this latter is a case for which we have not explicitly trained our features, and hence will test their generality.

\textbf{Partial-to-Full Shape Matching}
For partial-to-full shape matching, we use SOTA unsupervised methods, DPFM~\cite{attaiki2021dpfm} and ULRSSM~\cite{cao_unsupervised_2023}.
While originally they used spatial coordinates as input, this has been shown to heavily depend on the alignment of shapes in the datasets~\cite{ehm2025beyond, xie_echomatch_2025, attaiki2023generalizable}.
Therefore, we use vision foundation features as input.
We compare \methodName{} with \dinoThree~\cite{simeoni2025dinov3} features, as this latter is the second-best-performing (see \cref{sec:features_match}).
We report quantitative results for the partial-to-full matching in Table~\ref{tab:p2f_methods}, where the numbers in brackets show the features after test-time adaptation from ULRSSM.
We show a qualitative example in Figure~\ref{fig:p2f_shape_matching_qualitative}.
\begin{table}[tp!]
    \centering
    \small
    \begin{tabular}{lccc}
    \toprule
       Geo Err ($\downarrow$) & Features & ULRSSM~\cite{cao_unsupervised_2023} & DPFM~\cite{attaiki2021dpfm} \\
        \midrule
        \multirow{ 2}{*}{\becos{}} & \dinoThree{} & 26.97 (26.19) & 23.16\\
        & \textbf{\methodName} & \textbf{11.62 (11.54)} & \textbf{13.38}\\
        \midrule
        \multirow{ 2}{*}{\shrec{}\cuts{}} & \dinoThree{} & 5.94 (4.30) & 10.78\\
         & \textbf{\methodName} & \textbf{3.01 (1.97)} & \textbf{6.57}\\
        \midrule
        \multirow{ 2}{*}{\shrec{}\holes{}} & \dinoThree{} & 12.41 (9.66) & 10.41\\
         & \textbf{\methodName} & \textbf{8.07 (6.88)} & \textbf{7.38}\\
        \midrule
        \multirow{ 2}{*}{\pfaustM{}} & \dinoThree{} & 1.71 (1.69) & 3.78\\
         & \textbf{\methodName} & \textbf{1.61 (1.61)} & \textbf{3.62}\\
        \midrule
        \multirow{ 2}{*}{\pfaustH{}} & \dinoThree{} & 5.19 (5.22) & 6.46\\
         & \textbf{\methodName} & \textbf{2.29 (2.19)} & \textbf{5.66}\\
         \bottomrule
    \end{tabular}
    \caption{\textbf{Partial-to-Full Shape Matching with \methodName{}:} We show that we can improve the downstream task performance of partial-to-full shape matching with \methodName{} in comparison to \dinoThree{} features in terms of mean geodesic error ($\times 100$). The numbers in brackets of ULRSSM show the geodesic error after test-time adaptation.
    }
    \label{tab:p2f_methods}
\end{table}
For both methods, we observe the improvement provided by \methodName{} over the \dinoThree{} alternative.
Note that for the \becos{} dataset, the geodesic error is higher than when using \methodName{} directly.
We see two reasons for this: Firstly, \becos{} is the only dataset providing unnormalised shapes, meaning that the size ratio of the shapes is realistic (e.g., an elephant is larger than a mouse).
This may lead to incorrect predictions of the slanted diagonal ratio in the partial functional map prediction of ULRSSM and DPFM.
Secondly, the raycasting technique used to generate the partiality produces numerous boundaries, degrading the quality of the functional map.

\begin{table}[htp]
    \centering
    \small
    \begin{tabular}{lccccc}
    \toprule
        & Features & \multicolumn{2}{c}{EchoMatch~\cite{xie_echomatch_2025}} & \multicolumn{2}{c}{DPFM~\cite{attaiki2021dpfm}} \\
        \midrule
        && Err & mIoU & Err & mIoU \\
        \midrule
        \multirow{ 2}{*}{\becos{}} & \dinoThree{} & 9.74 & 67.07 & 9.83 & 49.80\\
        & \textbf{\methodName} & \textbf{5.55} & \textbf{71.27} & \textbf{8.20} & \textbf{56.04}\\
        \midrule
        \multirow{ 2}{*}{\psmal{}} & \dinoThree{} & 5.56 & 82.75  & 11.25 & 73.25\\
         & \textbf{\methodName}  & \textbf{4.33} & \textbf{85.41} & \textbf{8.91} & \textbf{78.95} \\
        \midrule
        \multirow{ 2}{*}{\cpTwopTwentyFour{}} & \dinoThree{} & \textbf{2.97} & 87.02 & 3.89 & \textbf{71.55} \\
         & \textbf{\methodName} & \textbf{2.97} & \textbf{87.92} & \textbf{3.48} & 70.41\\
        \bottomrule
    \end{tabular}
    \caption{\textbf{Partial-to-Partial Shape Matching with \methodName{}:} We demonstrate that our \methodName{} features achieve better downstream partial-to-partial shape matching performance than \dinoThree{}, as measured by geodesic error ($\times 100$) and Intersection over Union (IoU).}
    \label{tab:p2p_methods}
\end{table}
\newcommand{\sizeMichael}{0.25}
\newcommand{\spaceToShapes}{1cm}
\begin{figure}
    \centering
    \setlength{\tabcolsep}{2pt}
    \begin{tabular}{cccc}
    &Source & \dinoThree{} & \textbf{\methodName}\\
    \rotatebox{90}{\hspace{\spaceToShapes}DPFM}&\includegraphics[width=0.28\linewidth]{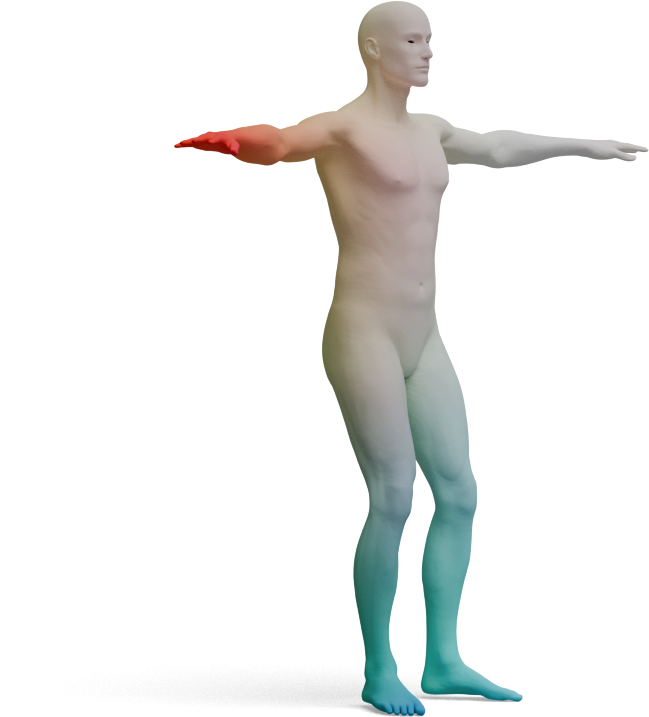}  & \includegraphics[width=\sizeMichael\linewidth]{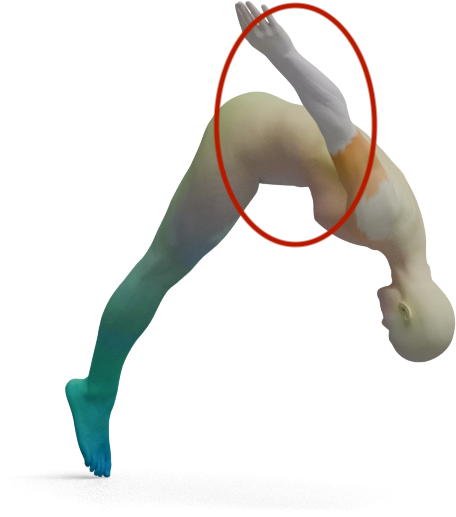} & \includegraphics[width=\sizeMichael\linewidth]{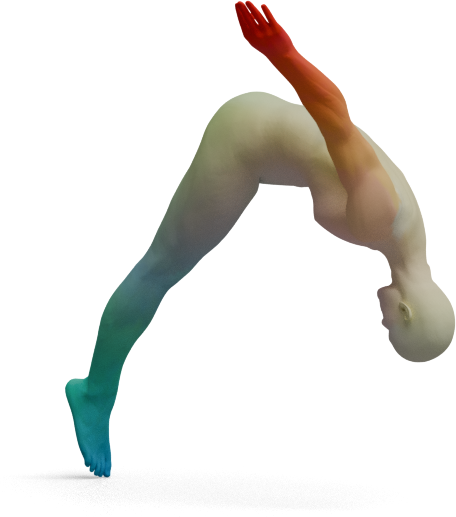}\\
    \rotatebox{90}{\hspace{\spaceToShapes}ULRSSM}&\includegraphics[width=0.28\linewidth]{p2f_results/pair_1/S1.png}  & \includegraphics[width=\sizeMichael\linewidth]{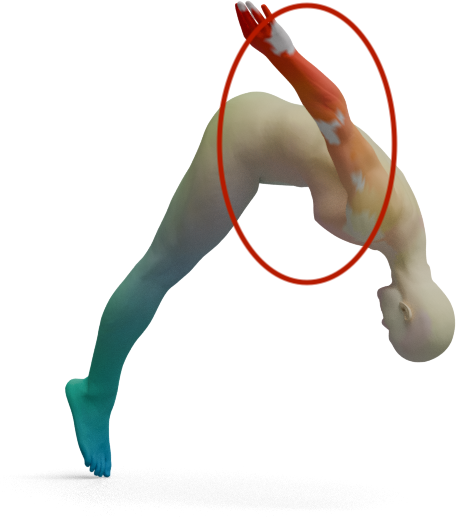} & \includegraphics[width=\sizeMichael\linewidth]{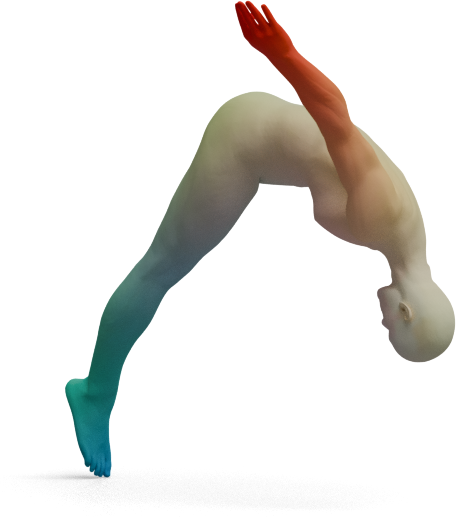}
    \end{tabular}
    
    \caption{\textbf{Qualitative Results of Partial-to-Full Shape Matching Methods:} We show one qualitative result of partial-to-full shape matching on the \shrec{} \cuts{} dataset.
    We can see improvement in the quality of the matching, especially at the arm with \methodName{} features.
    }
    \label{fig:p2f_shape_matching_qualitative}
\end{figure}

\textbf{Partial-to-Partial Shape Matching} Since no unsupervised methods are available for partial-to-partial shape matching, we rely on two SOTA supervised methods DPFM~\cite{attaiki2021dpfm} and EchoMatch~\cite{xie_echomatch_2025}.
In this setting, in conjunction with the correspondence, we also aim to identify the points that have a correspondence between the shapes, and thus determine their overlapping regions.
Following~\cite{ehm_partial--partial_2024}, we report both the geodesic error and the IoU in Table~\ref{tab:p2p_methods}.
For \becos{} and \psmal{}, we see improvement in terms of correspondence quality and overlap prediction.
For the \cpTwopTwentyFour{} dataset, we observe similar results with \methodName{} compared to \dinoThree{}, likely because the supervision prevents it from benefiting from improved feature descriptors.

\subsection{Ablation Studies}
To analyse our design choices, we train on the \becos{} full-to-full train set and validate on the \becos{} partial-to-full validation set.
Experiments were conducted on an NVIDIA A40 GPU and an Intel Xeon Gold 6248R CPU.
We show further ablation studies in the supplementary material.

\paragraph{Number of Views}
We present in Table~\ref{tab:num_views} an ablation study for the number of cameras/views $N$, reporting the mean geodesic error and runtime in seconds.
Using $49$ views is the sweet spot between the two metrics, as a larger number of views does not improve the mean geodesic error.
For fairness, we choose the same number of views for \methodName{} and the baselines \dinoTwo{} and \dinoThree{}.

\begin{table}[!tp]
    \centering
    \small
    \begin{tabular}{ccccccccc}
    \toprule
         &16&25&36&49&64&
         100\\
         \midrule
         Geo Err&
         25.47 & 20.80 &21.48&\textbf{20.09}	&20.29 & 
         \textbf{20.00}\\
        Time(s) & 
        \textbf{1.52} & 2.35 &3.36&4.59 &6.01&
        9.47\\
    \bottomrule
    \end{tabular}
    \caption{\textbf{Number of Views $N$ with Frozen \dinoThree{}:} With an increasing number of views, the time increases linearly, while the geodesic error ($\times 100$) already saturates earlier. Thus, we choose $49$ views for our experiments.}
    \label{tab:num_views}
\end{table}

\paragraph{Loss Formulation}
We compare our loss formulation with the widely used PointInfoNCE loss on the \becos{} validation set. 
We observe a strong improvement with \methodName{} in terms of geodesic error in Table~\ref{tab:lora}.
\begin{table}[htp]
    \centering
    \small
    \begin{tabular}{ccc}
    \toprule
         &PointInfoNCE& \textbf{\methodName}\\
         \midrule
         Err ($\downarrow$) &11.29& \textbf{5.89}\\
    \bottomrule
    \end{tabular}
    \caption{\textbf{PointInfoNCE vs \methodName:} We show that our loss formulation decreases the geodesic error ($\times 100$) significantly compared to the widely used PointInfoNCE loss.}
    \label{tab:lora}
\end{table}

\subsection{Applications}
\begin{figure}
    \centering
    \setlength{\tabcolsep}{0pt}
    \begin{tabular}{cccc}
        \rotatebox{90}{\hspace{0.5cm}DINOv3} & \includegraphics[width=0.4\linewidth]{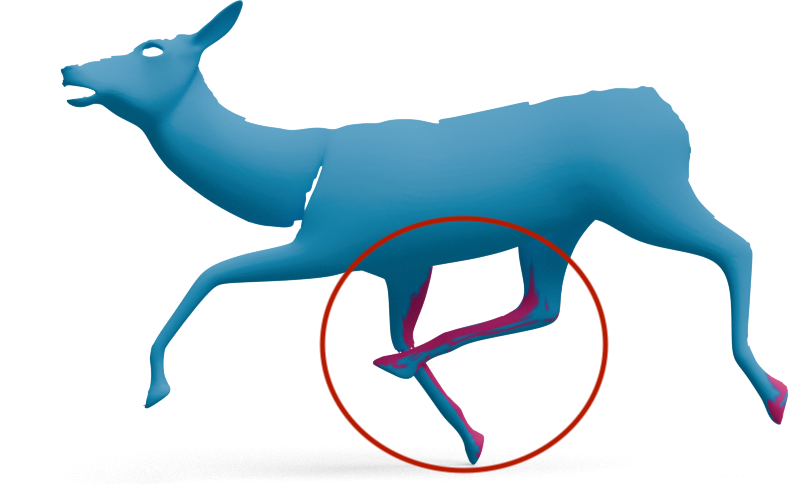}&
         \includegraphics[width=0.2\linewidth]{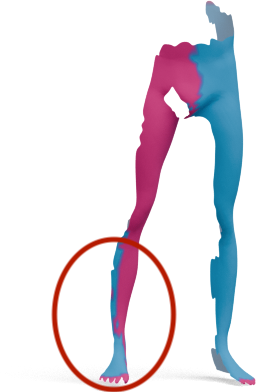} & 
         \includegraphics[width=0.27\linewidth]{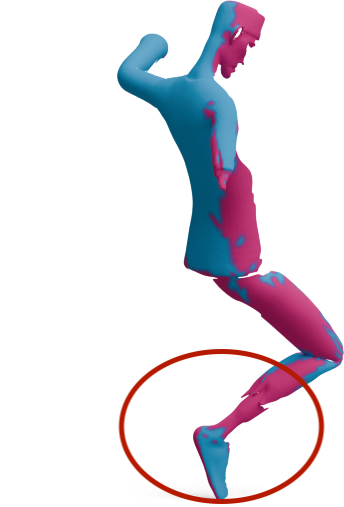}\\
        \rotatebox{90}{\hspace{0.5cm}\textbf{\methodName}} &  \includegraphics[width=0.4\linewidth]{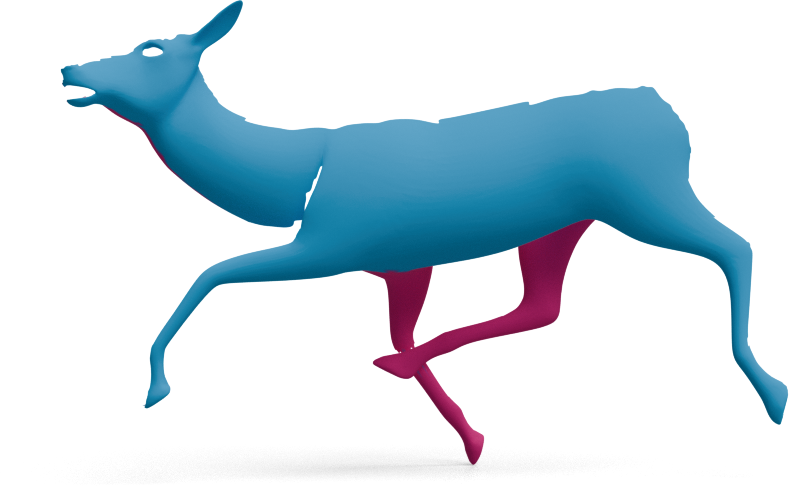} & 
         \includegraphics[width=0.2\linewidth]{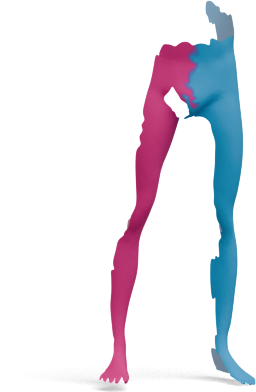} & 
         \includegraphics[width=0.27\linewidth]{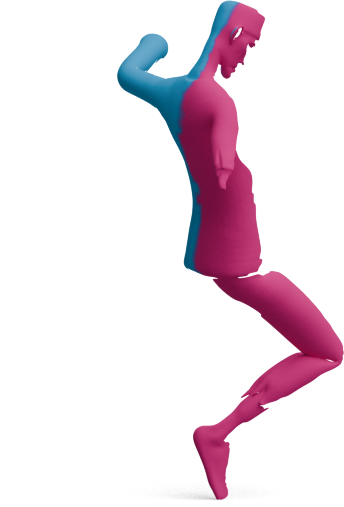}\\
    \end{tabular}
    
    \caption{\textbf{Qualitative Results of Left-Right Prediction:} \methodName{} shows better chirality prediction than DINOv3 (shapes from \becos{}).
    }
    \label{fig:qualitative_chirality}
\end{figure}

\paragraph{Chirality Classification}
Distinguishing the left and right of a mesh is an essential task for 3D shape analysis. 
We train the network from \cite{wang2025kh} with different features: \dinoThree{}, \methodName{}, and features obtained by our pipeline, but with the PointInfoNCE loss.
For training, we use the \becos{} full-to-full train set, and we evaluate on the partial shapes from the \becos{} partial-to-full test set. From the results in Table~\ref{tab:chirality}, we observe that \methodName{} provides significant improvement in terms of accuracy. The qualitative improvement can be seen in Figure~\ref{fig:qualitative_chirality}, even for highly partial shapes.
In comparison to the PointInfoNCE loss, our formulation further improves the left-right prediction.

\begin{table}[!tp]
    \centering
    \small
    \begin{tabular}{cccc}
    \toprule
    &\dinoThree{} & PointInfoNCE & \textbf{\methodName{}}\\
    \midrule
    Acc ($\uparrow$)& 70.45 & 84.09 & \textbf{91.42} \\
    \bottomrule
    \end{tabular}
    \caption{\textbf{Chirality Score for \becos{} Partial Shapes:} We show that we can improve the left-right prediction of~\cite{wang2025kh} with \methodName{} in terms of accuracy.
    }
    \label{tab:chirality}
\end{table}

\paragraph{Real-World Partiality} Mapping a partial observation to a full template is particularly relevant for real-world applications, such as interpreting data from acquisition pipelines.
In Figure~\ref{fig:real_world}, we report an illustrative result, obtained from a depth-sensor-based body scanner.
During the acquisition, the user was partially misaligned with respect to the capturing volume.
While \dinoThree{} fails to recover the semantics of such ruined geometry, our features show robustness. We found this result really exciting, as it can open up new techniques for non-rigid reconstruction of moving objects.

\newcommand{\templateHeight}{5.7cm}
\newcommand{\riccardoHeight}{6.2cm}
\begin{figure}
\setlength{\tabcolsep}{0pt}
    \centering
    \begin{tabular}{ccc}
       \includegraphics[width=1\linewidth, height=\templateHeight, keepaspectratio]{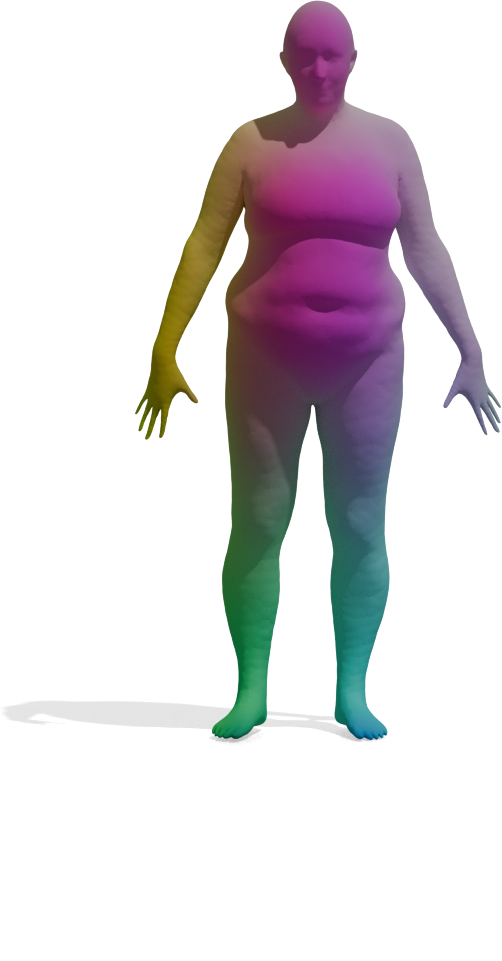}  & \includegraphics[width=1\linewidth, height=\riccardoHeight, keepaspectratio]{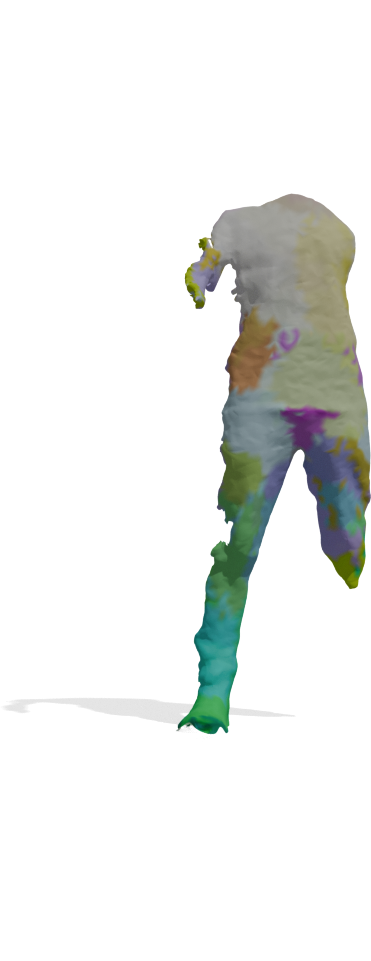} & \includegraphics[width=1\linewidth, height=\riccardoHeight, keepaspectratio]{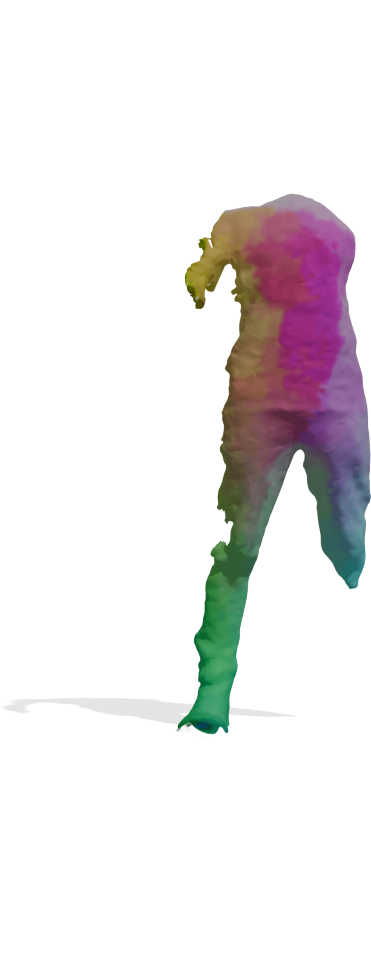} 
      \vspace{-1cm}
       \\
       \hspace{0.4cm}Template & \hspace{0.8cm}\dinoThree{} & \hspace{0.7cm}\textbf{\methodName}
    \end{tabular}
    \caption{\textbf{Real-World Scan:} We match a real-world partial scan on a template human. DINOv3 struggles to characterise incomplete geometry, while \methodName{} shows robust performance. }
    \label{fig:real_world}
\end{figure}

%% file: sec/5_conclusion.tex
\section{Discussion and Conclusion}
\label{sec:conclusion}
We introduced \methodName, a novel self-supervised approach for learning robust features on partial 3D shapes. By relying solely on synthetic data and self-supervised tasks, we demonstrated that our features not only possess a better geometrical understanding than 2D foundation backbones but are also the best available features to date for partial-to-partial and partial-to-full shape matching.
The main limitation is likely inherited from \dinoThree{}, which involves pretraining on natural images and thus incorporates a bias toward shapes in an upright orientation.
Although 3D objects are often assumed to be upright, this assumption might worsen the results in upside-down poses.
While our method still outperforms the \dinoThree{} baseline on upside-down poses, as shown in the supplementary material, both approaches show reduced performance compared to their results on upright-oriented shapes.
Overall, our work demonstrates that adapting vision foundation models through low-rank adaptation can substantially advance partial shape understanding.
In the future, our work could be extended to also consider external clutter and man-made objects.
Ultimately, we view our work as a promising step toward bridging the gap between large-scale image representations and 3D geometry, providing valuable insights to the broader communities of shape analysis and 3D vision.

%% file: sec/6_acknowledgments.tex
\section{Acknowledgments}
We thank Paul Roetzer for valuable feedback on our manuscript. This work was supported by the ERC Advanced Grant “SIMULACRON” (agreement \#884679), the ERC starting grant no. 101160648 (Harmony) and the Deutsche Forschungsgemeinschaft (DFG, German Research Foundation) – 458610525.

%% file: sec/X_suppl.tex
\clearpage
\setcounter{page}{1}
\maketitlesupplementary

\section*{Summary}
In this supplementary material, we provide additional details supporting the main paper.
We begin with an overview of the notation~(Sec.~\ref{sec:notation}) used throughout the method section, followed by extended ablation studies~(Sec.~\ref{sec:ablation}), like runtime comparisons, and a formal definition of the dataset-wide chirality accuracy~(Sec.~\ref{sec:chirality}).
We further describe implementation details~(Sec.~\ref{sec:additional-implementation-details}), compare our feature representation to those used in prior work (Sec.~\ref{sec:ulrssm-orig-features} and \ref{sec:chirality-original-features}), present failure cases~(Sec.~\ref{sec:failuer-cases}) and analyse the upright bias~(Sec.~\ref{sec:uprigth-bias}).
Additionally, we show more real-world scans~(Sec.~\ref{sec:real_world}) and provide additional qualitative results~(Sec.~\ref{sec:additional-qualitative-results}), including feature visualisations.
Finally, we present further 2D/3D application examples~(Sec.~\ref{sec:further2D_3D_application}) and we give some more insights on rendering with and without texture~(Sec.~\ref{sec:texture}).
\section{Notation Table}
\label{sec:notation}
For better understandability, we show an overview of the notation used in the method section in Table~\ref{tab:symbols}.
\begin{table}[htp]
    \centering
    \small
    \begin{tabular}{ll}
    \toprule
        Symbol & Description \\
    \midrule
       $\fullShape = (\vertFull, \facesFull)$ & 3D Full Shape\\
       $\vertFull$ & Vertices of shape $\fullShape$\\
       $\facesFull$ & Triangles of shape $\fullShape$\\
       $\partialShape = (\vertPartial, \facesPartial)$  & 3D Partial Shape, $\vertPartial \subset \vertFull$, $\facesPartial \subset \facesFull$ \\
       $\gtCorres$ & Correspondence from partial to full shape \\
       $\cameraSet = (\cameraFirst, ..., \cameraLast)$ & Set of all cameras\\      
       $\imageFull$ & Image of full shape from $\cameraIth \in \cameraSet$\\
       $\imageFeaturesFull$ & Features computed on $\imageFull$ \\
       $\shapeFeaturesFull$ & Vertex-wise features of shape $\fullShape$ \\
       $\geoMatrix$ & Geodesic Distance Matrix of $\fullShape$ \\
    \bottomrule
    \end{tabular}
    \caption{Overview of Symbols}
    \label{tab:symbols}
\end{table}

\section{Further Ablation Studies}
\label{sec:ablation}
\paragraph{LoRA Size}
We evaluate different values for hyperparameters, $\alpha$ and the rank $r$ in LoRA on the \becos{} validation set in terms of geodesic error (Table~\ref{tab:lora_size}).
The values $r=16$ and $\alpha=32$ yield the best results.

\begin{table}[htp]
    \centering
    \small
    \begin{tabular}{lccc}
    \toprule
    $r$/$\alpha$ &4/8& 8/16 & 16/32\\
     \midrule
     Geo Error ($\downarrow$) &8.14&6.56&\textbf{5.89}\\
    \bottomrule
    \end{tabular}
    \caption{\textbf{LoRA Hyperparameter Ablation:} We observe the best results with rank $r=16$ and $\alpha=32$.
    }
    \label{tab:lora_size}
\end{table}

\paragraph{Inference Time}
Table~\ref{tab:inference_time} compares the inference time of different methods.
The additional use of stable diffusion in Diff3f results in a longer processing time compared to using DINO features alone.
The other feature descriptors, including our method \methodName{}, have similar runtimes. 
\begin{table}[htp]
    \centering
    \small
    \begin{tabular}{ccccc}
    \toprule
    &  \dinoTwo{} & \dinoThree{} & Diff3f & \methodName{}\\
    \midrule
        Time [s] ($\downarrow$) & 4.72 & 4.45  & 360.45 & 4.59\\
    \bottomrule
    \end{tabular}
    \caption{\textbf{Inference Time Comparison:} While \dinoTwo{}, \dinoThree{} and \methodName{} have similar runtime during inference, Diff3f's runtime is significantly increased due to the need to additionally run inference with Stable Diffusion.}
    \label{tab:inference_time}
\end{table}

\paragraph{Backbone Scale}
Replacing the ViT-B backbone with a ViT-L backbone does not lead to significant performance gains in terms of geodesic error (see Table~\ref{tab:backbone}).
Nevertheless, ViT-L requires 50\% more training time and twice the amount of VRAM on the GPU than ViT-B.
\begin{table}[htp]
\centering
\footnotesize
\footnotesize
\begin{tabular}{ccccc}
\toprule
   PFAUST & DINOv3-B & \methodName{}-B & DINOv3-L & \methodName{}-L\\
   \midrule
    -M & 11.36 & 2.88 & 11.40 & \textbf{2.43}\\
    -H & 13.82 & 3.33 & 12.50 & \textbf{2.65}\\
\bottomrule
\end{tabular}
\caption{\textbf{Performance of Different Backbone Scales:} We show that with a larger backbone we do not gain significant improvement in terms of geodesic error.}
    \label{tab:backbone}
\end{table}

\section{Dataset-wide Chirality Accuracy Definition}
\label{sec:chirality}
Given a full dataset $U$ with $M$ partial shapes $U=\{\partialShape_1, ..., \partialShape_M\}$, we define the chirality accuracy as:
\begin{equation}
    acc_U = \frac{1}{M} \sum_{\partialShape_i \in U} \acc_{\partialShape_i}
\end{equation}
We can define the final chirality accuracy as:
\begin{equation}
    acc_{\ciralitySym} = \text{max}\{acc_U, 1- acc_U\}
\end{equation}.

\section{Additional Implementation Details} \label{sec:additional-implementation-details}
For the temperature parameter, we follow~\cite{attaiki2021dpfm} and use $\tau = 0.07$.
We run the training on one A40 GPU and five cores of a Xeon Gold 6248R CPU.
The training takes approximately 5 days for $50,000$ iterations for the longest-lasting \becos{} dataset.
The training time can vary between 2-5 days between the different datasets according to different number of vertices.
As a prompt to Stable Diffusion inside Diff3f, we use "animal" for animal shapes, "human" for human shapes and "centaur" for centaur shapes.

\section{ULRSSM: Comparison to Original Features}\label{sec:ulrssm-orig-features}
Instead of using foundation features or handcrafted descriptors, an alternative approach is to start with the raw XYZ coordinates as input. Since previous methods have shown promising results, for fairness, we also compare our approach with ULRSSM~\cite{cao_unsupervised_2023} using XYZ coordinates as input. In Table~\ref{tab:original_features}, we show that the performance falls far behind the use of foundation features.
The use of XYZ is sensitive to shape orientations, which are particularly ambiguous in partial shapes.
\begin{table}[htp]
    \small
    \centering
    \begin{tabular}{lcc}
    \toprule
        Geo Error $(\downarrow)$ & XYZ & Ours\\
        \midrule 
        \becos{} & 40.46 (40.18) &\textbf{11.62 (11.54)}\\        
        \shrec{}\cuts{} & 7.04 (3.2) &\textbf{3.01 (1.97)}\\
        \shrec{} \holes{} & 13.75 (8.2) &\textbf{8.07 (6.88)} \\
        \pfaustM{} & 7.65 (7.49) &\textbf{1.61 (1.61)} \\
        \pfaustH{} & 9.39 (9.24) & \textbf{2.29 (2.19)}\\
         \bottomrule
    \end{tabular}
    \caption{\textbf{Comparison of XYZ features vs \methodName{} features:} We compare the geodesic error $(\times 100)$ on different partial-to-full datasets with ULRSSM. On all datasets, our features outperform the originally used XYZ features. The numbers in brackets ``$()$'' indicate geodesic error results after test-time adaptation~\cite{cao_unsupervised_2023}.} 
    \label{tab:original_features}
\end{table}

\section{Left-Right Prediction: Comparison to Original Features}\label{sec:chirality-original-features}
In the main paper, we compare the \methodName{} features with those of \dinoThree{}. 
For completeness, here we report a comparison with the features originally used in \cite{wang2025kh}, which combine \dinoTwo{} and Stable Diffusion features (Table~\ref{tab:supp_chirality}).
\begin{table}[htp]
\centering
    \begin{tabular}{ccc}
    \toprule
         & Original & Ours\\
         \midrule
       Chirality Accuracy ($\uparrow$)  & 57.40 & \textbf{91.42}\\
         \bottomrule
    \end{tabular}
    \caption{\textbf{Comparison to DINOv2 + Stable Diffusion Features:} We compare the chirality (left-right) accuracy of the features used in \cite{wang2025kh} to our \methodName{} features. 
    Our method's features show higher accuracy.}
    \label{tab:supp_chirality}
\end{table}

\section{Failure Cases}\label{sec:failuer-cases}
\begin{figure}[htp]
    \centering
      \setlength{\tabcolsep}{-6pt}
      \begin{tabular}{cccc}
    \includegraphics[width=0.3\linewidth]{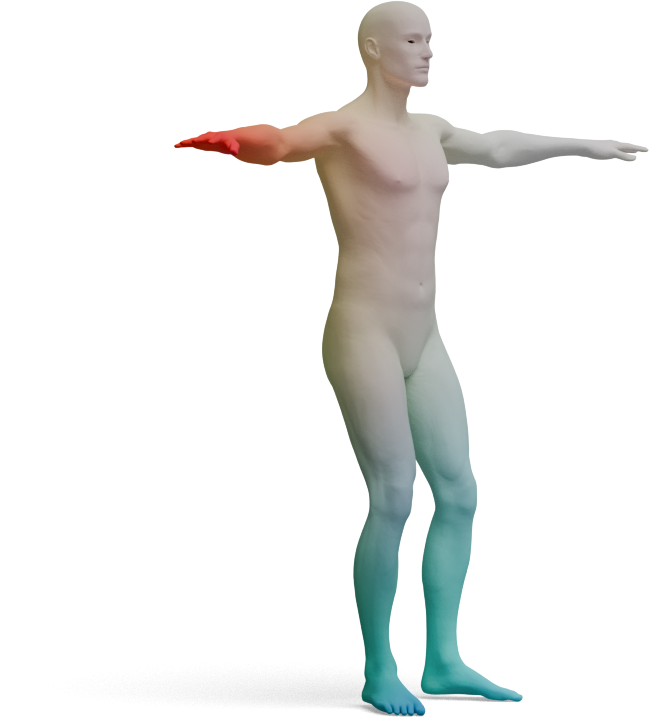} & 
    \includegraphics[width=0.28\linewidth]{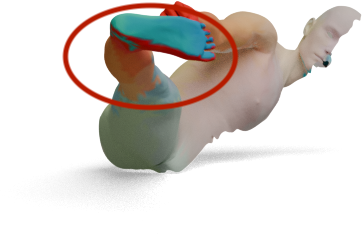}  & 
    \includegraphics[width=0.3\linewidth]{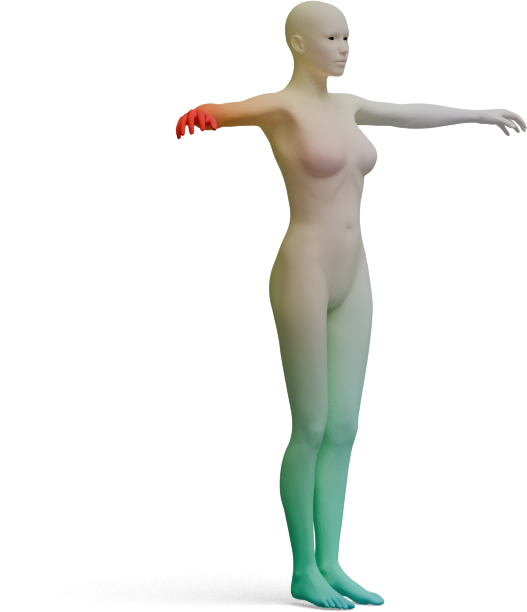}  & 
    \includegraphics[width=0.28\linewidth]{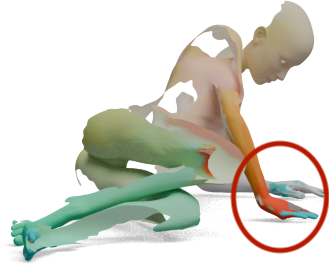} 
    \end{tabular}
    \caption{\textbf{Failure Cases:} We show two failure cases of our method. The method primarily struggles with shapes in challenging poses, where the shapes are not positioned upright.}
    \label{fig:failure_cases}
\end{figure}

The datasets used to train our models primarily contain upright shapes, resulting in most failure cases involving shapes in different poses, such as a rotated person in a pike position or a person lying on their side with their hands on the ground.
We show two examples in Figure~\ref{fig:failure_cases}.

\begin{figure}[htp]
  \centering
  \begin{tabular}{ccc}
    \multirow[t]{4}{*}{%
      \hspace{-2cm} 
      \raisebox{-2.7cm}{
      \includegraphics[width=0.4\linewidth]{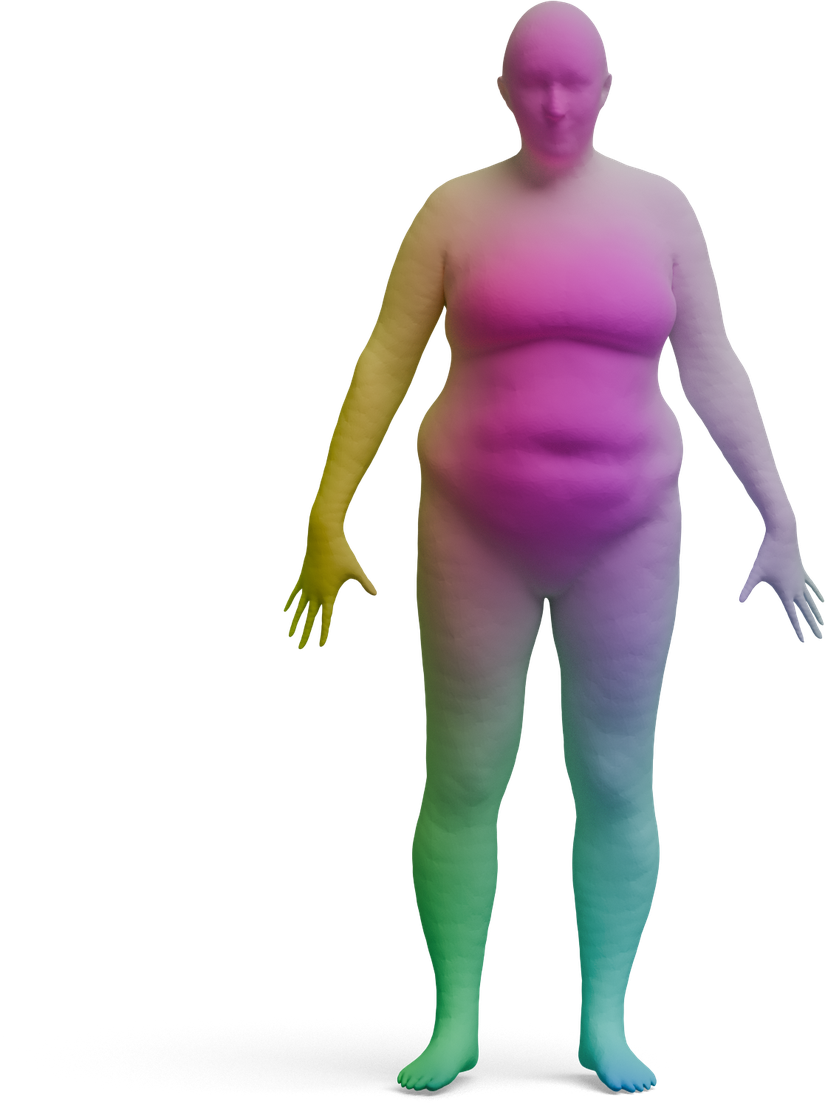}
      }
    }
    & \includegraphics[width=0.2\linewidth]{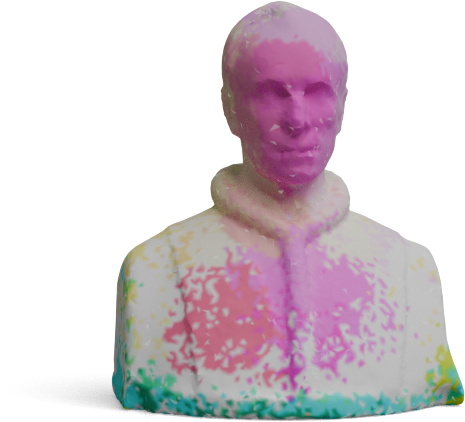}
    & \includegraphics[width=0.2\linewidth]{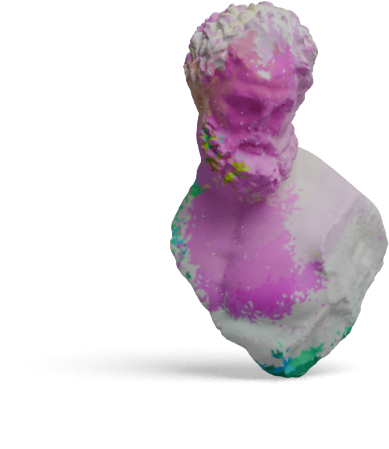} \\
    & \multicolumn{2}{c}{\dinoThree{}} \\ 
    & \includegraphics[width=0.2\linewidth]{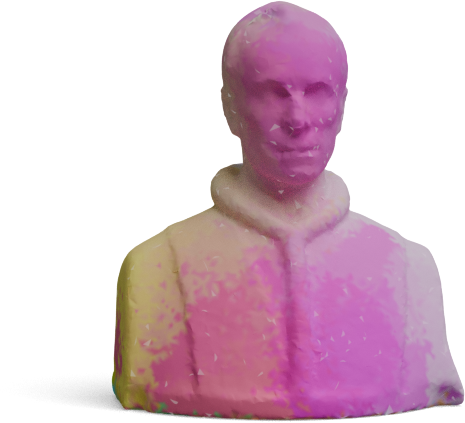}
    & \includegraphics[width=0.2\linewidth]{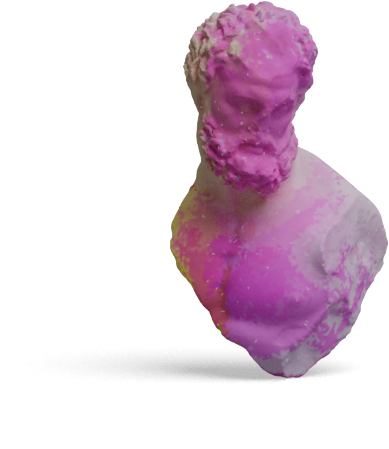} \\
    & \multicolumn{2}{c}{\textbf{Ours}} \\
  \end{tabular}
\caption{\textbf{Additional Results on Real-World Statues from \cite{ScanTheWorldMMF}:} \dinoThree{} fails to find good correspondences between the partial statues and the full template shape, where \methodName{} finds reasonably dense correspondences.}
\end{figure}

\section{Upright Bias}\label{sec:uprigth-bias}
For consistency with the literature, we assume that shapes are oriented upright. 
However, we train \methodName{} on FAUST (for 70K iterations), sample random rotation along all axes, and test upside-down partial shapes. \methodName{} achieves 6.69 geodesic error, while DINOv3 performs 10$\times$ worse (63.84).

\section{Additional Real-World Experiments}\label{sec:real_world}
\paragraph{Additional Real World Scans of Statues}
We demonstrate further real-world scans of statues from~\cite{ScanTheWorldMMF} and match them to a template shape from \faust{}.
Here, \methodName{} shows qualitatively better matchings than \dinoThree{}.

\paragraph{Real-World Validation}
Quantitative evaluation on real-world partial scans is limited by the lack of ground-truth data availability.
Therefore we evaluate the annotated real scan of SHREC20.
Our method significantly reduces geodesic error compared to DINOv3 (5.92 vs. 16.26), indicating improved robustness on real-world data.\\

\section{Additional Qualitative Results}\label{sec:additional-qualitative-results}

\paragraph{Feature Quality} We show more qualitative comparisons to other features in Figure~\ref{fig:supp_qualitative_p2f}.
Additionally, we show further examples of our feature predictions in Figure~\ref{fig:supp_qualitative_ours_more}.

\paragraph{Partial Shape Matching}
We show qualitative results for partial-to-partial shape matching in Figure~\ref{fig:supp_qualitative_p2p} and additional results for partial-to-full shape matching in Figure~\ref{fig:supp_p2f_matching}.

\section{Further 2D/3D Applications}\label{sec:further2D_3D_application}
\begin{figure}[htp]
    \centering
    \begin{tabular}{cc@{\hspace{0.1cm}}c@{\hspace{0.1cm}}c}
    \hspace{-0.15cm}\rotatebox[origin=l]{90}{\hspace{-0.1cm}\footnotesize{DINOv3}} & 
    \includegraphics[width=0.315\linewidth]{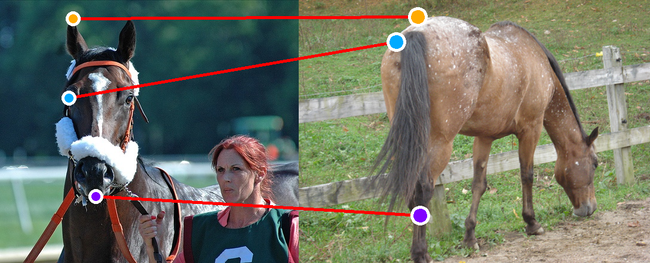} & \includegraphics[width=0.37\linewidth]{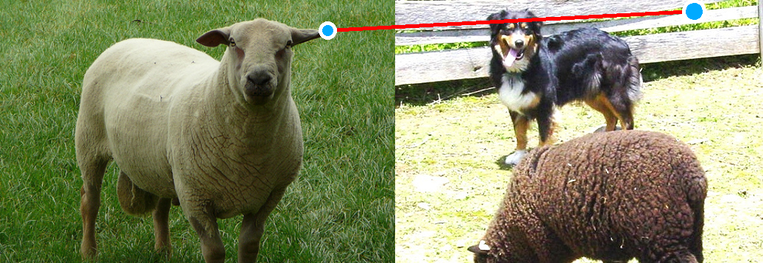}& \includegraphics[width=0.18\linewidth]{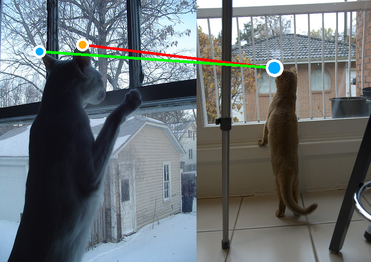}\\
    \hspace{-0.2cm}
    \rotatebox[origin=l]{90}{\hspace{0.2cm}\footnotesize{Ours}} & \includegraphics[width=0.315\linewidth]{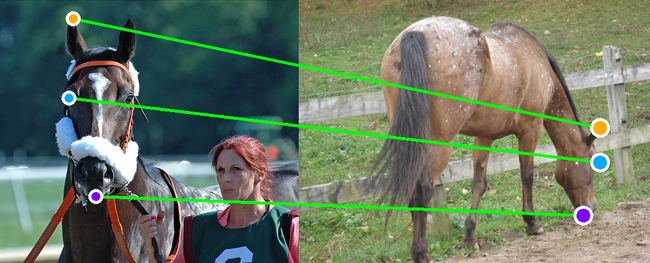} & \includegraphics[width=0.37\linewidth]{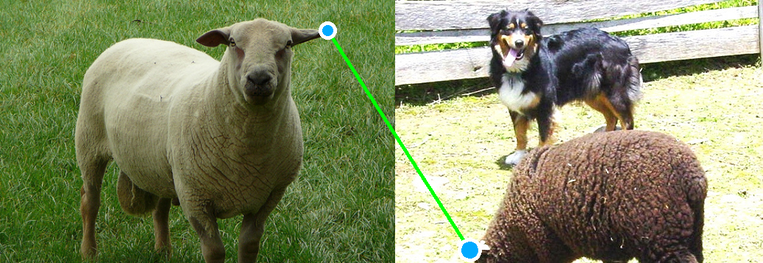}& \includegraphics[width=0.18\linewidth]{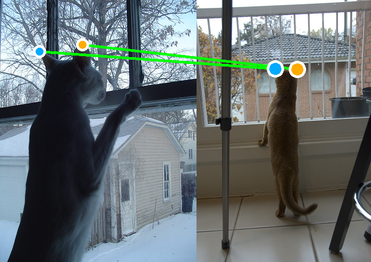}\\
\end{tabular}
    \caption{\textbf{2D Keypoint Matching:} We show examples for keypoint matching on the SPair-71k~\cite{min2019spair} dataset. Our method shows superior matching results in these examples.}
    \label{fig:keypoints2D}
\end{figure}

\begin{figure}[htp]
\centering
\begin{tabular}{cc|cc}
\includegraphics[width=0.17\linewidth]{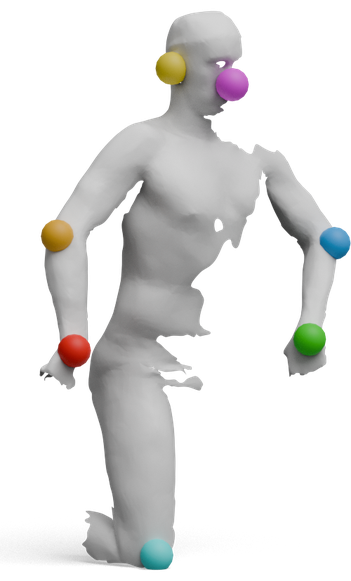}  & 
\hspace{-0.75cm}
\includegraphics[width=0.22\linewidth]{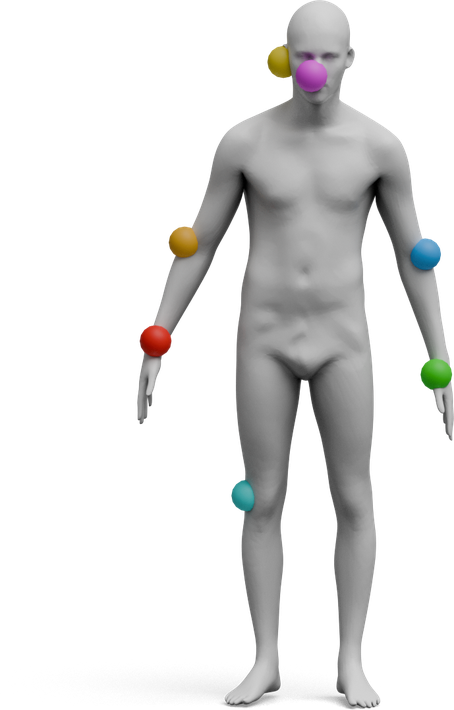} 
&
\includegraphics[width=0.23\linewidth]{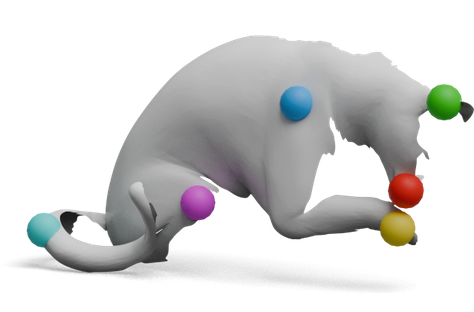}
& 
\hspace{-0.7cm}
\includegraphics[width=0.33\linewidth]{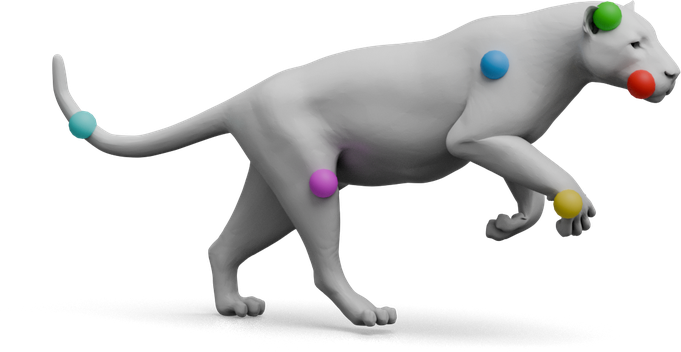}
\\
\end{tabular}
    \caption{\textbf{3D Keypoint Matching:} Our method can be used for 3D keypoint matching. We show examples of the BeCoS test set.}
    \label{fig:keypoints3D}
\end{figure}

While we focus on learning features for 3D partial shape matching, we find their use in other contexts exciting. 
Hence, we use our \methodName{} model (trained on BeCoS) to perform image matching on SPair-71k~\cite{min2019spair} dataset images.
In several cases, our approach provides better matching than DINOv3, especially in images where foreground elements differ in their view~(see Table~\ref{fig:keypoints2D}).
Additionally, we show that our method can also be used for 3D keypoint matching for shapes of the BeCoS test set~(see Table~\ref{fig:keypoints3D}).

\section{Texture \& Rendering}\label{sec:texture}
All partial shape matching benchmarks are textureless, so meshes are rendered without texture (single colour with shadow on the surface, see Fig.~2 Image Column).
Using SMPL~\cite{loper2015smpl}, we empirically test texture sensitivity and find that adding texture degrades performance (geodesic error 3.65 vs. 1.32).
We hypothesise that texture helps identify correspondences within the same (e.g., person), but not across different identities (e.g., due to different clothing).\\

\newcommand{\cutsSourceOne}{0.15}
\newcommand{\cutsOne}{0.12}
\newcommand{\holesSourceTwo}{0.15}
\newcommand{\holesTwo}{0.10}
\newcommand{\pfaustMSource}{0.15}
\newcommand{\pfaustMThree}{0.15}
\newcommand{\cutsFigSupp}[3]{\includegraphics[width=#2\linewidth]{figures/supp_features_comparison/pair_#3/#1.png}}

\begin{figure*}[htp]
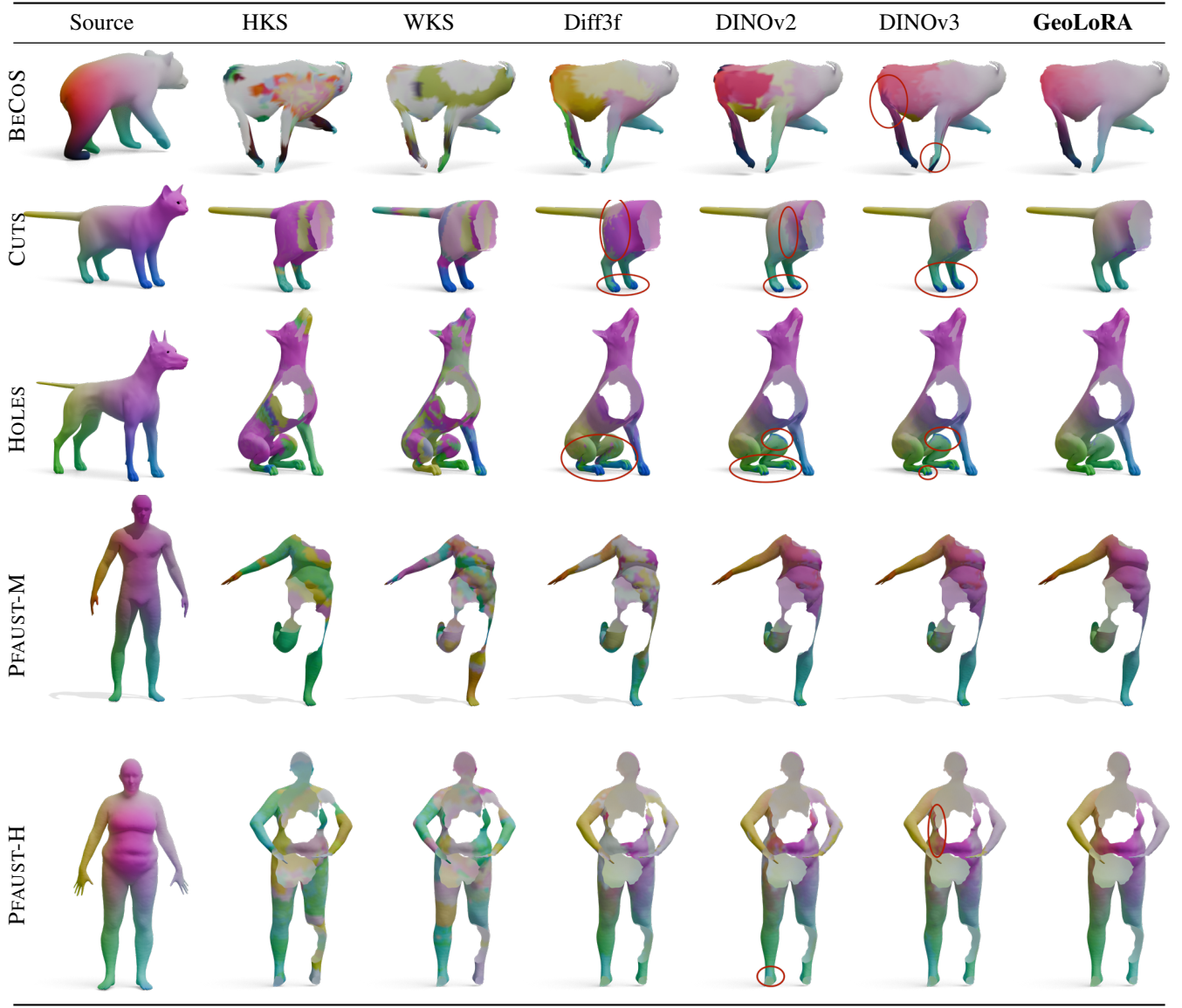

\centering
\setlength{\tabcolsep}{-2pt}
\begin{tabular}{lccccccc}
\toprule
  & Source & HKS & WKS & Diff3f & \dinoTwo{} & \dinoThree{} & \textbf{\methodName} \\
  \midrule
    \rotatebox{90}{\hspace{0.5cm}\becos{}} &
    \cutsFigSupp{S1}{\pfaustMSource}{0} &
    \cutsFigSupp{S2_hks}{\pfaustMThree}{0}  &
    \cutsFigSupp{S2_wks}{\pfaustMThree}{0}  &
    \cutsFigSupp{S2_diff3f}{\pfaustMThree}{0}  & 
    \cutsFigSupp{S2_dinoV2}{\pfaustMThree}{0}  &
    \cutsFigSupp{with_circles/S2_dinoV3}{\pfaustMThree}{0}  &
    \cutsFigSupp{S2_ours}{\pfaustMThree}{0}  \\
     \rotatebox{90}{\hspace{0.5cm}\cuts{}} &
     \cutsFigSupp{S1}{\cutsSourceOne}{1} &
         \cutsFigSupp{S2_hks}{\cutsOne}{1} &
     \cutsFigSupp{S2_wks}{\cutsOne}{1} & 
     \cutsFigSupp{with_circles/S2_diff3f}{\cutsOne}{1} & 
     \cutsFigSupp{with_circles/S2_dinoV2}{\cutsOne}{1} & 
     \cutsFigSupp{with_circles/S2_dinoV3}{\cutsOne}{1} & 
     \cutsFigSupp{S2_ours}{\cutsOne}{1} \\
    \rotatebox{90}{\hspace{0.5cm}\holes{}} &
    \cutsFigSupp{S1}{\holesSourceTwo}{2} &
    \cutsFigSupp{S2_hks}{\holesTwo}{2} &
    \cutsFigSupp{S2_wks}{\holesTwo}{2} &
    \cutsFigSupp{with_circles/S2_diff3f}{\holesTwo}{2} &
    \cutsFigSupp{with_circles/S2_dinoV2}{\holesTwo}{2} &
    \cutsFigSupp{with_circles/S2_dinoV3}{\holesTwo}{2} &
    \cutsFigSupp{S2_ours}{\holesTwo}{2} \\
    \rotatebox{90}{\hspace{1cm}\pfaustM{}} &
    \cutsFigSupp{S1}{\pfaustMSource}{3} &
    \cutsFigSupp{S2_hks}{\pfaustMThree}{3} &
    \cutsFigSupp{S2_wks}{\pfaustMThree}{3} &
    \cutsFigSupp{S2_diff3f}{\pfaustMThree}{3} & 
    \cutsFigSupp{with_circles/S2_dinoV2}{\pfaustMThree}{3} &
    \cutsFigSupp{with_circles/S2_dinoV3}{\pfaustMThree}{3} &
    \cutsFigSupp{S2_ours}{\pfaustMThree}{3} \\
    \rotatebox{90}{\hspace{1cm}\pfaustH{}} &
    \cutsFigSupp{S1}{\pfaustMSource}{4} &
    \cutsFigSupp{S2_hks}{\pfaustMThree}{4} &
    \cutsFigSupp{S2_wks}{\pfaustMThree}{4} &
    \cutsFigSupp{S2_diff3f}{\pfaustMThree}{4} & 
    \cutsFigSupp{with_circles/S2_dinoV2}{\pfaustMThree}{4} &
    \cutsFigSupp{with_circles/S2_dinoV3}{\pfaustMThree}{4} &
    \cutsFigSupp{S2_ours}{\pfaustMThree}{4} \\
   \bottomrule
\end{tabular}
\caption{\textbf{Qualitative Results of Feature Quality:} We show colour-coded correspondences based on the features computed with different feature extractors. \methodName{} exhibits improved performance, especially in left-right and front-back predictions.
}
\label{fig:supp_qualitative_p2f}
\end{figure*}

\begin{figure*}[htp]
    \centering
  \setlength{\tabcolsep}{4pt}
    \begin{tabular}{cc:cc:cc}
\includegraphics[width=0.15\textwidth]{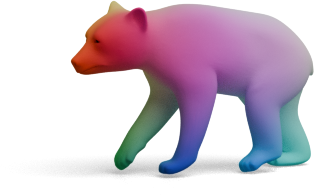} &
\includegraphics[width=0.15\textwidth]{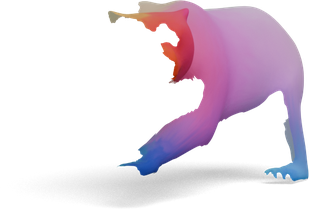} &
\includegraphics[width=0.15\textwidth]{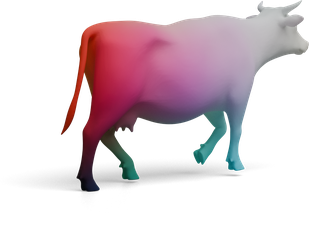} &
\includegraphics[width=0.15\textwidth]{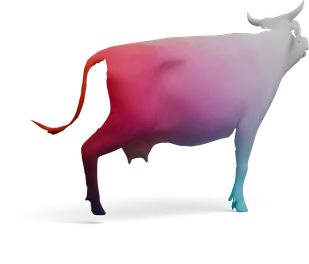} &
\includegraphics[width=0.15\textwidth]{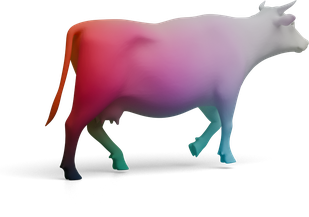} &
\includegraphics[width=0.15\textwidth]{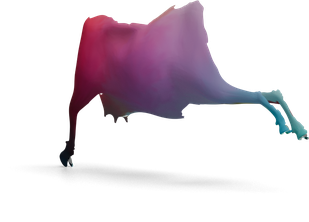} \\
\includegraphics[width=0.15\textwidth]{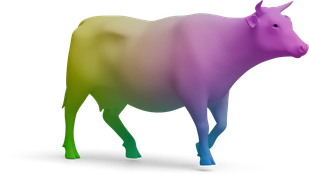} &
\includegraphics[width=0.15\textwidth]{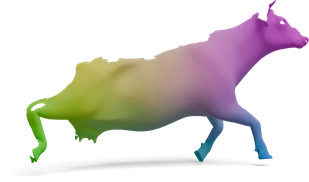} &
\includegraphics[width=0.15\textwidth]{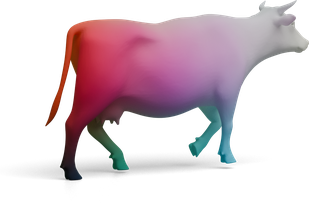} &
\includegraphics[width=0.12\textwidth]{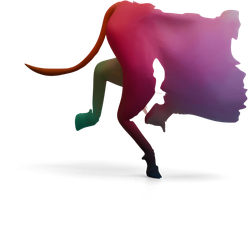} &
\includegraphics[width=0.12\textwidth]{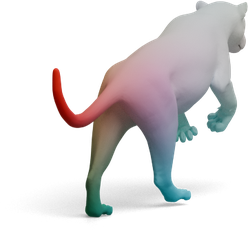} &
\includegraphics[width=0.12\textwidth]{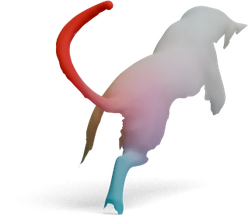} \\
\includegraphics[width=0.15\textwidth]{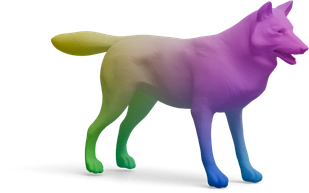} &
\includegraphics[width=0.10\textwidth]{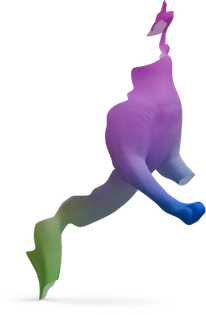} &
\includegraphics[width=0.15\textwidth]{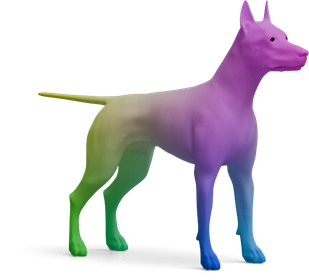} &
\includegraphics[width=0.15\textwidth]{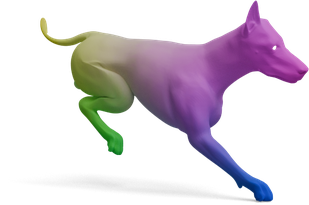} &
\includegraphics[width=0.15\textwidth]{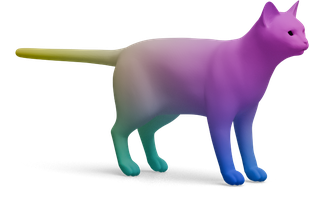} &
\includegraphics[width=0.11\textwidth]{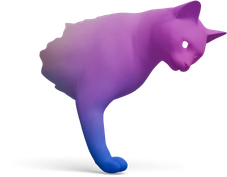} \\
\includegraphics[width=0.15\textwidth]{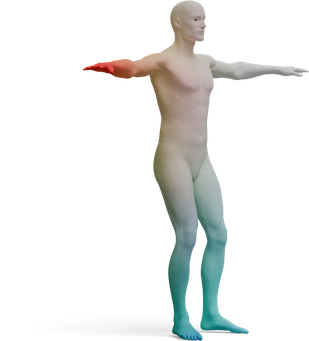} &
\includegraphics[width=0.11\textwidth]{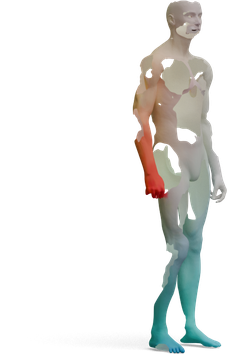} &
\includegraphics[width=0.15\textwidth]{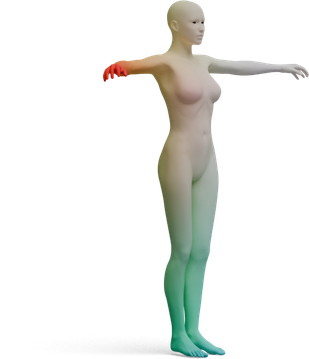} &
\includegraphics[width=0.12\textwidth]{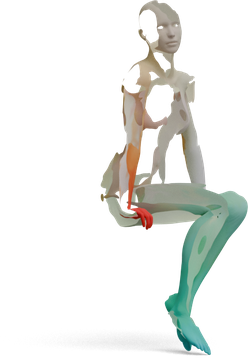} &
\includegraphics[width=0.15\textwidth]{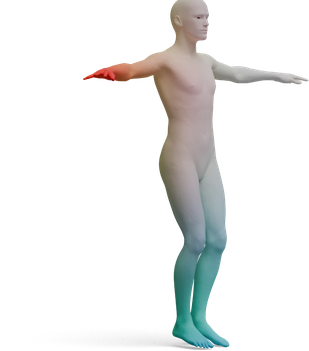} &
\includegraphics[width=0.12\textwidth]{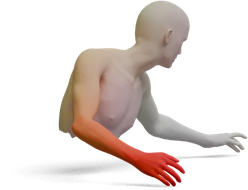} \\
\includegraphics[width=0.15\textwidth]{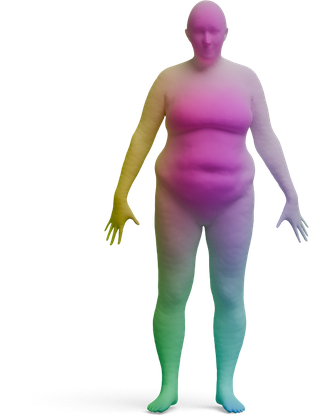} &
\includegraphics[width=0.12\textwidth]{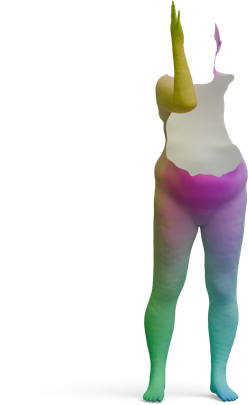} &
\includegraphics[width=0.15\textwidth]{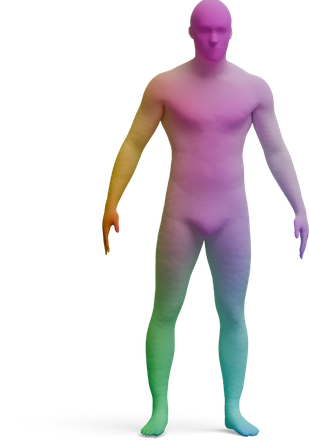} &
\includegraphics[width=0.15\textwidth]{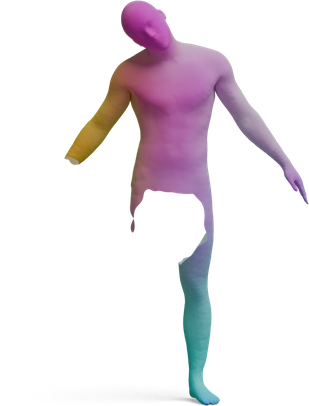} &
\includegraphics[width=0.15\textwidth]{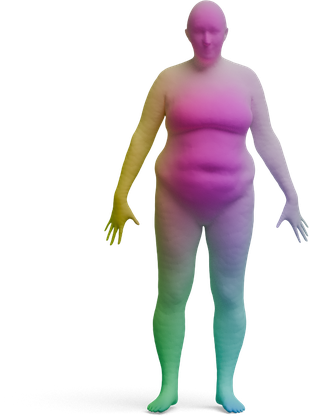} &
\includegraphics[width=0.12\textwidth]{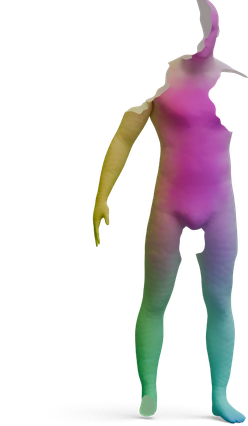} \\
\includegraphics[width=0.12\textwidth]{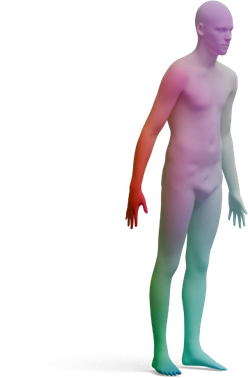} &
\includegraphics[width=0.07\textwidth]{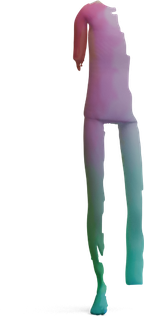} &
\includegraphics[width=0.15\textwidth]{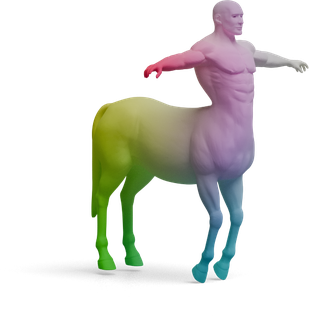} &
\includegraphics[width=0.12\textwidth]{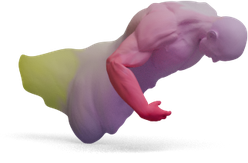} &
\includegraphics[width=0.15\textwidth]{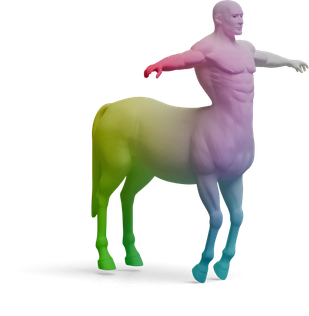} &
\includegraphics[width=0.15\textwidth]{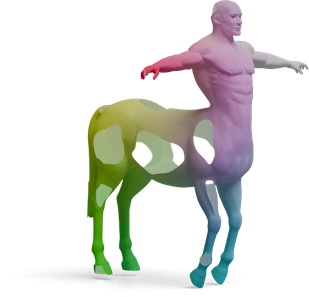}
\\
\end{tabular}
    \caption{\textbf{More Qualitative Results:} We show more qualitative results of our partial-to-full features.}
    \label{tab:more_qualitative_ours}
\label{fig:supp_qualitative_ours_more}
\end{figure*}

\begin{figure*}[htp!]
\centering
\begin{tabular}{cccccc}
\toprule
    &Source & \multicolumn{2}{c}{DPFM} & \multicolumn{2}{c}{EchoMatch} \\
    \midrule
     && \dinoThree{} & \textbf{\methodName{}}& \dinoThree{} & \textbf{\methodName{}} \\
    \midrule
    \multirow[t]{2}{*}{\rotatebox{90}{\becos{}}}&\includegraphics[width=0.13\textwidth]{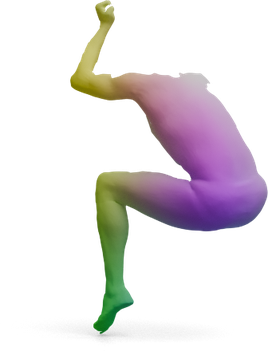} &
\includegraphics[width=0.12\textwidth]{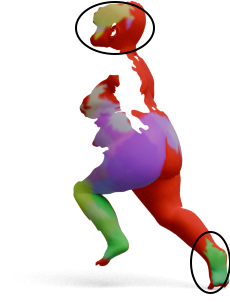} &
\includegraphics[width=0.12\textwidth]{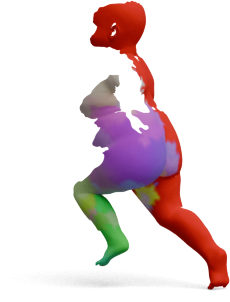} &
\includegraphics[width=0.12\textwidth]{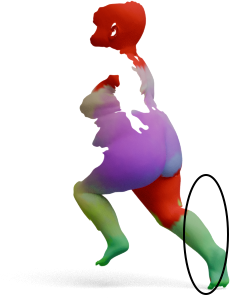} &
\includegraphics[width=0.12\textwidth]{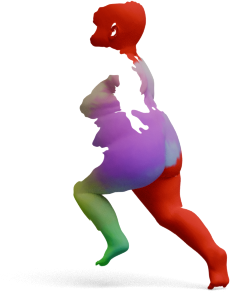} \\
     & \includegraphics[width=0.2\textwidth]{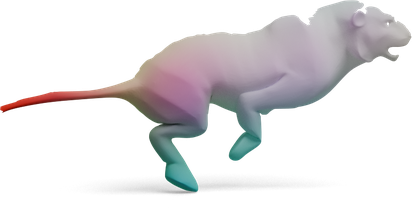} &
\includegraphics[width=0.15\textwidth]{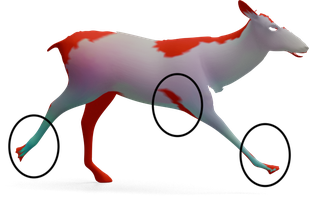} &
\includegraphics[width=0.15\textwidth]{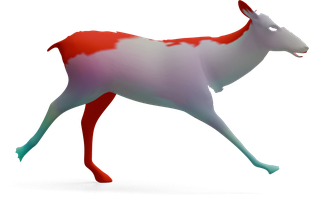} &
\includegraphics[width=0.15\textwidth]{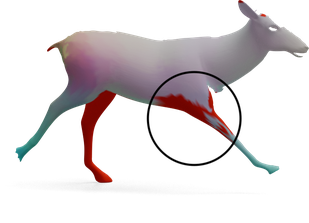} &
\includegraphics[width=0.15\textwidth]{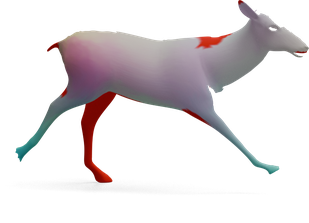} \\
\midrule

\multirow[c]{2}{*}{\rotatebox{90}{\psmal{}}} & \includegraphics[width=0.12\textwidth]{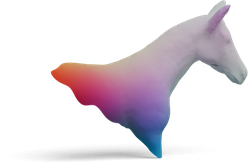} &
\includegraphics[width=0.15\textwidth]{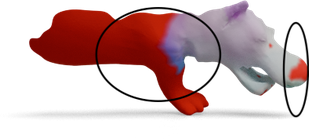} &
\includegraphics[width=0.15\textwidth]{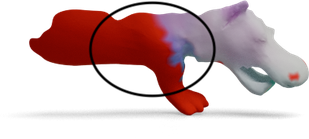} &
\includegraphics[width=0.15\textwidth]{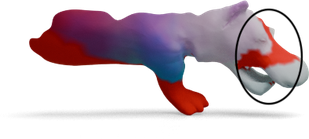} &
\includegraphics[width=0.15\textwidth]{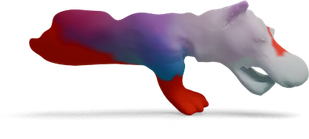} \\
& \includegraphics[width=0.16\textwidth]{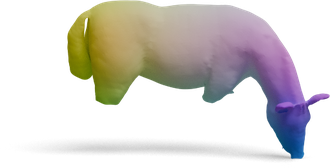} &
\includegraphics[width=0.15\textwidth]{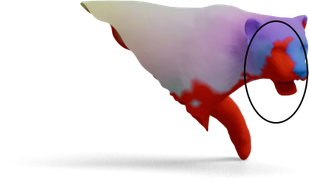} &
\includegraphics[width=0.15\textwidth]{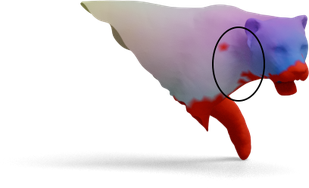} &
\includegraphics[width=0.15\textwidth]{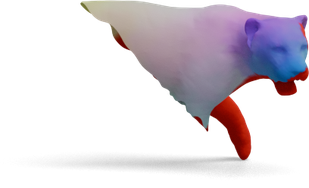} &
\includegraphics[width=0.15\textwidth]{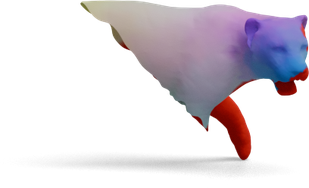} \\
\midrule
 \multirow[c]{2}{*}{\rotatebox{90}{\hspace{-1.3cm}\cpTwopTwentyFour{}}}& \includegraphics[width=0.08\textwidth]{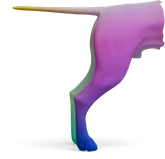} &
\includegraphics[width=0.15\textwidth]{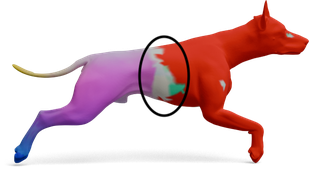} &
\includegraphics[width=0.15\textwidth]{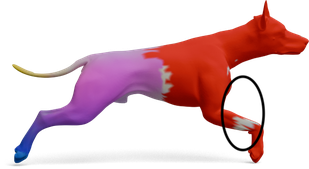} &
\includegraphics[width=0.15\textwidth]{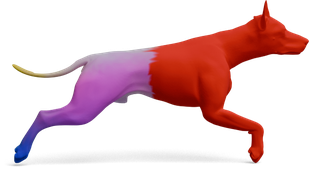} &
\includegraphics[width=0.15\textwidth]{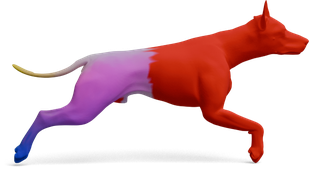} \\
& \includegraphics[width=0.1\textwidth]{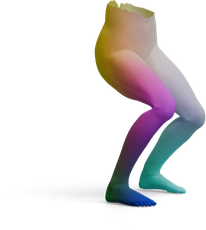} &
\includegraphics[width=0.12\textwidth]{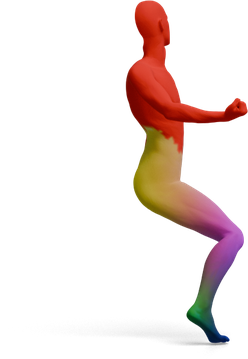} &
\includegraphics[width=0.12\textwidth]{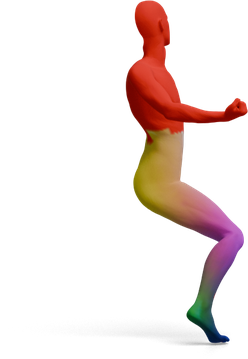} &
\includegraphics[width=0.12\textwidth]{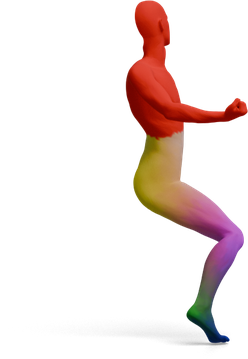} &
\includegraphics[width=0.12\textwidth]{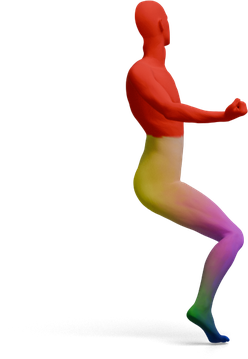} \\
\bottomrule
\end{tabular}
\caption{\textbf{Qualitative Results of Partial-to-Partial Matching:} We show colour-coded correspondences for partial-to-partial shape matching methods. We compare \dinoThree{} features with \methodName{} features. The red area indicates points that are predicted not to be in the overlapping region.
On the \becos{} dataset and the \psmal{} dataset, we demonstrate improved performance with \methodName{} features compared to \dinoThree{} features. 
On \cpTwopTwentyFour{} we observe similar matching quality.
}
\label{fig:supp_qualitative_p2p}
\end{figure*}

\newcommand{\sizeMichaelSupp}{0.15}
\newcommand{\sizeHorseSupp}{0.15}

\begin{figure*}[htp]
    \centering
    \setlength{\tabcolsep}{2pt}
    \begin{tabular}{cccccc}
    \toprule
    &Source & \multicolumn{2}{c}{DPFM} & \multicolumn{2}{c}{ULRSSM} \\
    \midrule
    & & \dinoThree{} & \textbf{\methodName}& \dinoThree{} & \textbf{\methodName}\\
    \midrule
    \rotatebox{90}{\hspace{1.5cm}\pfaustH{}} & \includegraphics[width=0.17\linewidth]{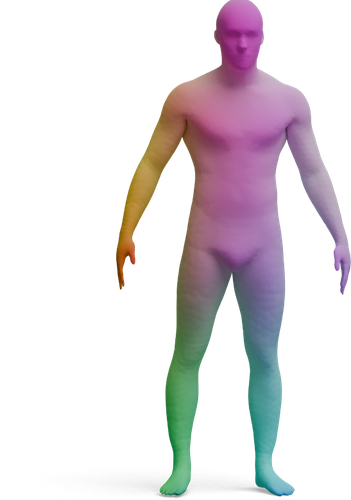}  & \includegraphics[width=\sizeMichaelSupp\linewidth]{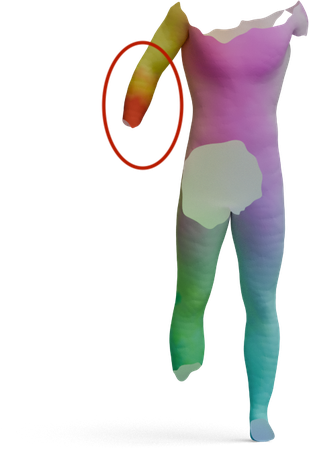} & \includegraphics[width=\sizeMichaelSupp\linewidth]{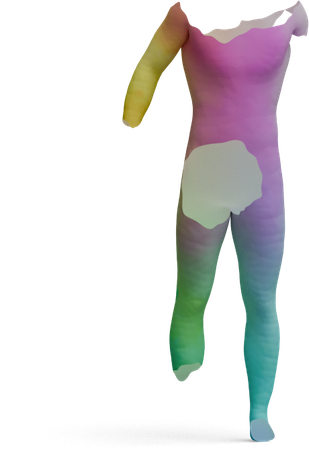}
    & \includegraphics[width=\sizeMichaelSupp\linewidth]{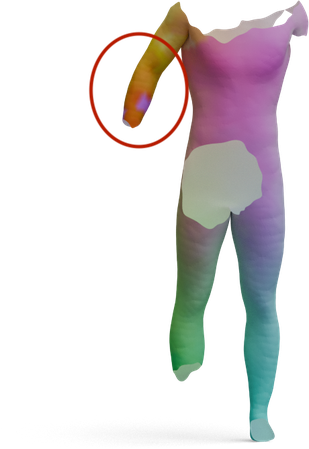} & \includegraphics[width=\sizeMichaelSupp\linewidth]{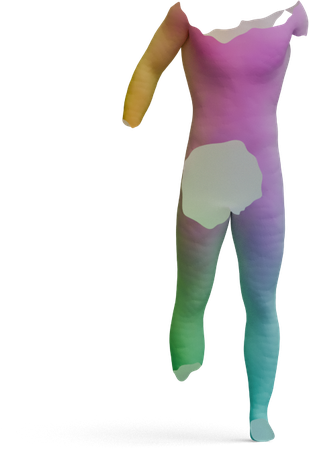} \\
    \midrule
    \rotatebox{90}{\hspace{0.8cm}\holes{}} & \includegraphics[width=0.24\linewidth]{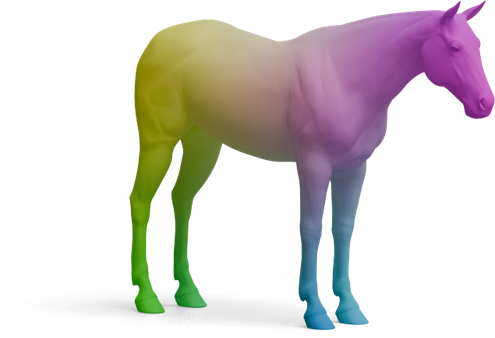}  & \includegraphics[width=\sizeHorseSupp\linewidth]{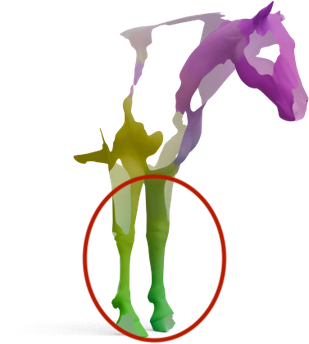} & \includegraphics[width=\sizeHorseSupp\linewidth]{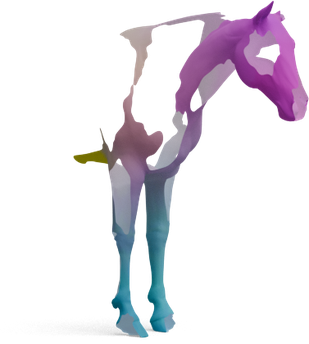}  & \includegraphics[width=\sizeHorseSupp\linewidth]{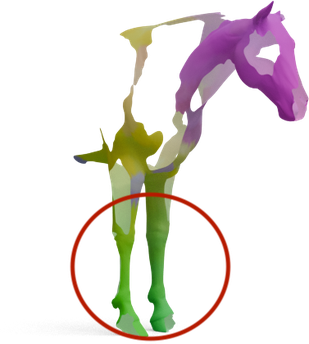} & \includegraphics[width=\sizeHorseSupp\linewidth]{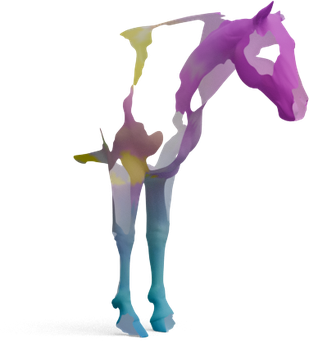}\\
    \bottomrule
    \end{tabular}
    
    \caption{\textbf{Qualitative Results of Partial-to-Full Shape Matching Methods:} We show two more examples of partial-to-full shape matching on the \pfaustH{} and \shrec{} \holes{} dataset.
    We can see improved matching quality with our \methodName{} features, especially in the extremities of the shapes, such as the left arm of the human and the legs of the horse.
    }
    \label{fig:supp_p2f_matching}
\end{figure*}